%% file: acl_latex.tex
\documentclass[11pt]{article}

\usepackage[final]{acl}

\usepackage{times}
\usepackage{latexsym}

\usepackage[T1]{fontenc}

\usepackage[utf8]{inputenc}

\usepackage{microtype}

\usepackage{inconsolata}

\usepackage{graphicx}

\usepackage[inline]{enumitem}
\usepackage{booktabs, multirow, tabularx}
\usepackage[table]{xcolor}
\usepackage{amsmath}
\usepackage{amssymb}
\usepackage{listings,float,caption,subcaption}
\usepackage{makecell}
\usepackage{fontawesome5}

\newcommand{\eg}{\textit{e.g.,} }
\newcommand{\ie}{\textit{i.e.,} }
\definecolor{jbcolor}{RGB}{180, 50, 120}

\definecolor{promptbg}{RGB}{248,248,248}
\definecolor{promptframe}{RGB}{180,180,180}
\definecolor{keywordcolor}{RGB}{0,90,160}

\lstdefinestyle{promptstyle}{
  backgroundcolor=\color{promptbg},
  basicstyle=\ttfamily\scriptsize,
  breaklines=true,
  breakatwhitespace=false,
  frame=single,
  rulecolor=\color{promptframe},
  xleftmargin=4pt,
  xrightmargin=4pt,
  aboveskip=2pt,
  belowskip=2pt,
  columns=fullflexible,
  keepspaces=true,
  showstringspaces=false,
}

\title{Clarify, Abstain or Answer?\\Strategising in Conversation with Belief-Augmented Generation}

\author{Joris Baan\textsuperscript{\faBicycle}, Wilker Aziz\textsuperscript{\faBicycle}, Barbara Plank\textsuperscript{\faMountain\faCar}, Raquel Fern{\'a}ndez\textsuperscript{\faGlobe }\textsuperscript{\faBicycle} \\
        {\footnotesize \faBicycle}University of Amsterdam, 
        {\scriptsize \faMountain}MCML Munich,
        {\scriptsize \faCar}LMU Munich, \\ 
        {\scriptsize \faGlobe }European Commission, Joint Research Centre (JRC), Brussels
        \\
        \texttt{\{j.s.baan,w.aziz,raquel.fernandez\}@uva.nl},  \texttt{b.plank@lmu.de}}

\begin{document}
\maketitle

\input{sections/0-abstract}
\input{sections/1-introduction}
\input{sections/2-related_work}
\input{sections/3-method}
\input{sections/4-experimental_setup}
\input{sections/5-evaluation}

\input{sections/6-results-direct}
\input{sections/7-results-bag}

\input{sections/8-faithfulness}

\input{sections/10-conclusion}
\input{sections/11-acknowledgements}

\input{sections/12-limitations}

\bibliography{anthology-1,anthology-2, custom}
\input{sections/appendix}

\end{document}

%% file: sections/0-abstract.tex
\begin{abstract}

Large language models (LLMs) define a distribution over text, which can be viewed as a probabilistic representation of uncertainty: sampling $K$ responses yields a \textit{belief state}---responses a model deems plausible. 
Existing work exploits this representation for narrow tasks like either decoding or selective prediction, and often requires manual interventions, not controlling generation directly. 
We propose Belief-Augmented Generation (BAG): grounding LLMs in their own belief state via the prompt and letting them reason over these $K$ samples to decide on and execute a conversational strategy: clarify, abstain, or answer. 
In a multi-turn ambiguous question answering (QA) setting, we find that LLMs by default rarely clarify or abstain, ignoring uncertainty about the input (aleatoric) or facts (epistemic). 
BAG improves QA accuracy across six models and yields strategy decisions more faithful to their belief state than prompt-only baselines. 
Disentangling when to clarify from when to abstain, however, remains challenging.\footnote{Code available at \href{https://github.com/jsbaan/belief-augmented-generation}{github.com/jsbaan/belief-augmented-generation}, and an explorer to browse questions, generations and evaluations at \href{https://jorisbaan.nl/belief-augmented-generation}{jorisbaan.nl/belief-augmented-generation}.}

\end{abstract}

%% file: sections/1-introduction.tex
\section{Introduction}

Large language models (LLMs) are statistical models that autoregressively define a distribution over text. This can be viewed as a probabilistic representation of uncertainty or belief \cite{baan2023uncertainty}. By drawing multiple samples we obtain a textual representation, \ie \textit{belief state}: responses a model deems plausible under a given prompt. 

Such representations are routinely and effectively used in various research subfields: in decoding, by selecting the highest expected utility or majority generation \cite{eikema-aziz-2020-map, wang2023selfconsistency}; in selective prediction, abstaining when entropy across semantic clusters of generations is high \cite{kuhn2023semantic}; in clarification finetuning, to synthesise clarification questions \cite{zhang2025modeling}; and in verbalised uncertainty, to communicate uncertainty reflective of the relative frequency of a claim in the belief state \cite{yona-etal-2024-large, eikema2025teaching}.

Our main contribution is to unify these disparate research strands, all using the same set of samples for a single narrow decision-making task, and do so without  manual heuristics and interventions (\eg decoders, uncertainty quantifiers, selective prediction heuristics, or finetuning). 

\begin{figure}[t]
    \centering
    \includegraphics[width=0.99\linewidth]{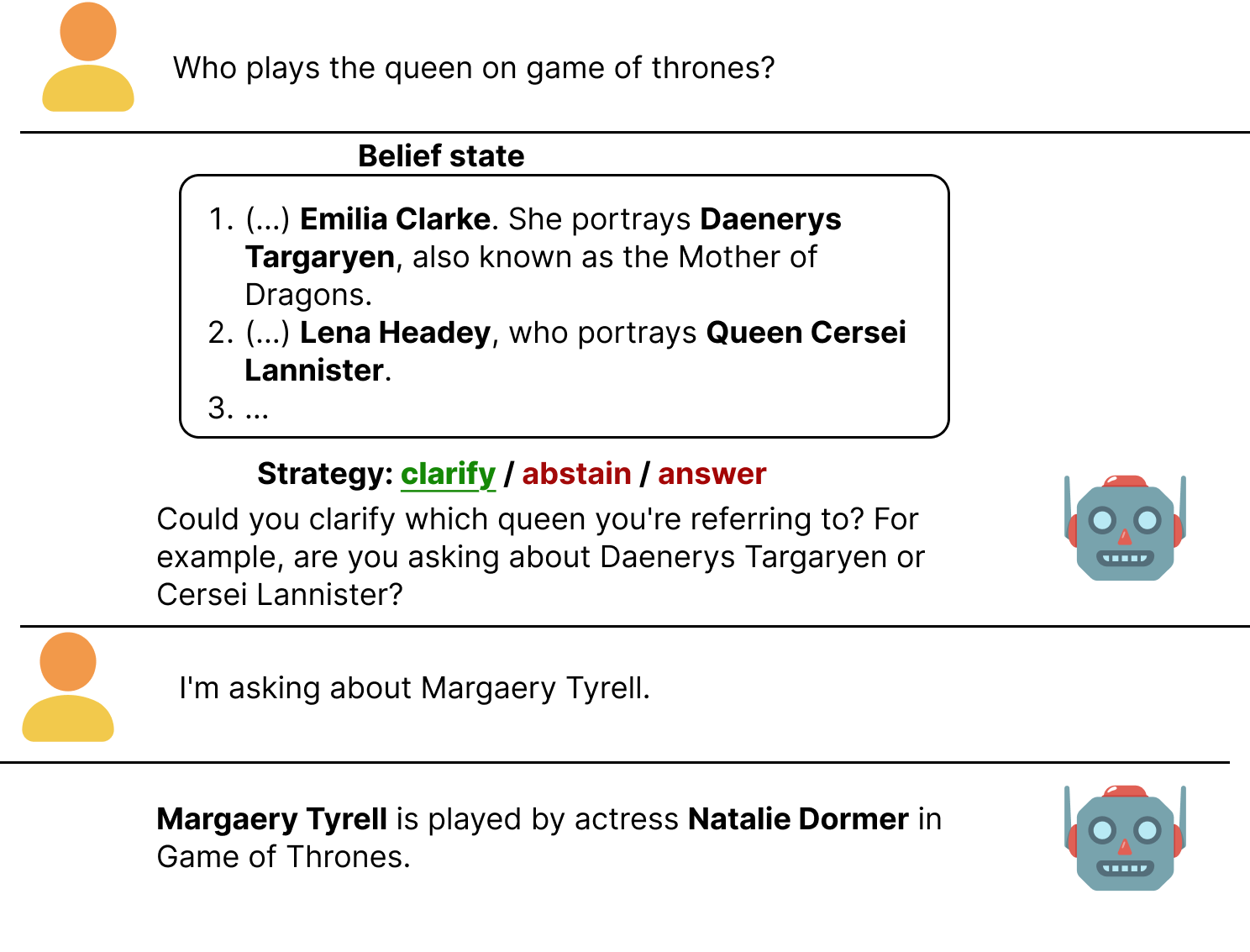}
    \caption{\textbf{Turn 1}: User asks a potentially ambiguous question. \textbf{Turn 2}: BAG samples $K$ responses, analyses them, and formulates a strategy and response. \textbf{Turn 3}: BAG asked a clarification question so we simulate a user answer. \textbf{Turn 4}: Final answer, optionally with another round of BAG. Real Qwen3-14B example.}
    \label{fig1:bag}
\end{figure}

We capitalise on the intuition that inspecting multiple generations provides information about reliability. For example, sample diversity can be characteristic of different sources of uncertainty and, as such, informative of when and how to clarify, abstain, or answer. Specifically, we let LLMs reason over an explicit textual representation of their own uncertainty, obtained via sampling, guiding them to clarify, abstain or answer directly. This harnesses LLMs' strong abilities to reason about available textual evidence, as demonstrated in retrieval-augmented generation or RAG \citep{lewis2020retrieval}. 

We call our method \textit{Belief-Augmented Generation} or BAG: conditioning generation directly on the belief state (rather than external documents in RAG) via the prompt. BAG lets an LLM decide when to answer, what strategy to follow, and how to formulate the answer all at once. BAG can be viewed as test-time compute scaling \cite{snell2025scaling}, requiring no training, no external data, and being applicable to any model, including closed models. Figure \ref{fig1:bag} shows an example. 

The six LLMs in our experiments by default rarely clarify or abstain and ignore uncertainty in the input (sometimes called aleatoric or data uncertainty, for example due to ambiguity) and about facts (sometimes called epistemic or model uncertainty). In contrast, belief-augmented generation improves QA performance and leads to strategies more faithful to probabilistic uncertainty.

To complement automatic QA evaluation of increasingly complex and verbose LLM generations---which is non-trivial \cite{kamalloo-etal-2023-evaluating,adlakha-etal-2024-evaluating}---we also release \href{https://jorisbaan.nl/belief-augmented-generation/}{BAG Explorer \faDesktop }, an online tool to manually browse the questions, references, generations and evaluations used in this paper. 

%% file: sections/2-related_work.tex
\section{Related Work}

\paragraph{Decoding.}
One of the earlier works that exploit a textual representation of probabilistic belief (\ie $K$ sampled generations) is Minimum Bayes Risk decoding (MBR), which selects the generation that best represents the entire set, demonstrating improved response quality \cite{eikema-aziz-2020-map,suzgun-etal-2023-follow,wu2025better}. 
Similarly, self-consistency samples multiple reasoning chains and pick the majority answer \cite{wang2023selfconsistency,bertsch-etal-2023-mbr}. Both can be viewed as test-time compute scaling \cite{snell2025scaling}.

\paragraph{Selective prediction.} 
\citet{kuhn2023semantic, farquhar2024detecting} instead capitalise on the idea that semantically distinct claims in the belief state may be indicative of error (\ie model or epistemic uncertainty)---an intuition widely used to detect hallucinations \cite{geng-etal-2024-survey, vashurin-etal-2025-benchmarking}. Others view variation in belief states as uncertainty about the input (\citealp[\ie data or aleatoric uncertainty; ][]{giulianelli-etal-2023-comes,baan-etal-2024-interpreting}), and there have been efforts to disentangle these sources \cite{cole-etal-2023-selectively, hou-etal-2024-decomposing}. However, both decoding and selective prediction methods rely on hand-crafted features and their end goal is to predict error or select a generation rather than directly controlling generation.

\paragraph{Faithful uncertainty.}
Another line of research guides models to communicate their uncertainty faithfully, \ie hedge or strengthen a claim such that it reflects the relative frequency of that claim in the belief state \cite{yona-etal-2024-large, xu-etal-2024-sayself, eikema2025teaching}. Most similar to our work is \citet{kirchhof2025self}, who propose to communicate the full internal distribution over possible answers rather than hedging a single answer. They find that directly conditioning on $K$ samples in the prompt is most effective. We go beyond a simple summary and instead instruct models to analyse their own belief state to power dynamic conversational decisions. Also similar is \citet{nan-etal-2025-multiple} who condition a bigger, auxiliary model on another model's belief state to analyse the source of its uncertainty. We instead let the same model reason about its own uncertainty to directly inform its generation process.

\paragraph{Conversational strategies.}
Predicting conversational strategies has a long history in (spoken) dialogue systems, from decision-theoretic action selection \citep{horvitz2000conversation,horvitz2000deeplistener} to belief representation and estimation \citep{bohus2005constructing,bohus2005principled,williams-etal-2013-dialog}. 

However, LLMs use a limited range of conversational strategies and often answer directly, failing to recognize ambiguity or ask clarification questions, instead implicitly assuming an interpretation \cite{deng-etal-2023-prompting, shaikh-etal-2024-grounding, zhang-etal-2024-clamber, testoni-etal-2025-racquet}. While there have been attempts to remedy this via prompting \cite{kuhn2022clam,zhang-etal-2024-clamber} and fine-tuning \cite{andukuri2024stargate,chen2025learning}, the belief state has been under-explored.

\citet{testoni-fernandez-2024-asking} suggest that human clarification questions as supervision for deciding when and how to clarify may not be ideal to resolve model uncertainty, which is echoed by \citet{kim-etal-2024-aligning} who finetune on questions `perceived' by a model as ambiguous, and \citet{zhang2025modeling} who finetune on synthetic clarifications generated by an external model instructed to distinguish between $K$ sampled generations. \citet{zhang-choi-2025-clarify} predict ambiguity via entropy over multiple simulated user responses without exploiting them for how to clarify, and do not allow abstaining.
 
Crucially, unlike classical selective prediction, BAG guides models to abstain with a direct response to the user based on the belief state (see the example in Figure \ref{fig:bag_3examples}). In fact, the abstain strategy can easily be replaced by or enhanced with faithfully communicating uncertainty, or fetching external information, \ie deciding when to RAG. 
Similarly, clarification is but one strategy to address ambiguity. There are others, like long form answers \cite{berant2026learning, saparina2026reasoning}, which are fully compatible with BAG. 
We are not aware of recent work combining direct answer, clarification and abstention strategies.

%% file: sections/3-method.tex
\begin{figure*}
    \centering
    \includegraphics[width=0.99\linewidth]{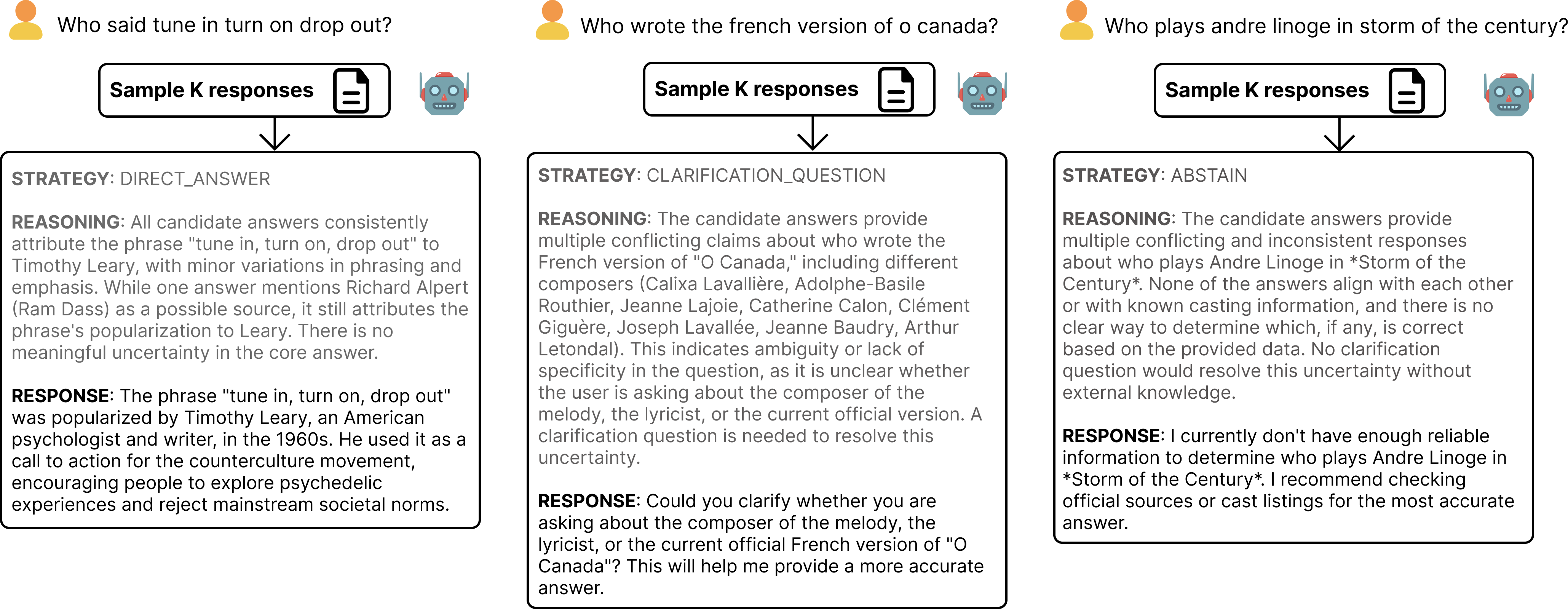}
    \caption{Three real examples of Qwen3-14B output when strategising about the best conversational strategy based on the user's question and its own belief state (which we omit for brevity) using BAG2. User only sees the \textbf{response}.}
    \label{fig:bag_3examples}
\end{figure*}

\section{Belief-Augmented Generation (BAG)}
\label{sec:bag}

BAG feeds a textual representation of probabilistic belief directly into the prompt, guiding models to improve their responses by making belief-informed conversational decisions. We capitalise on the intuition that inspecting multiple generations gives insight into the response reliability, \eg that sample diversity can be characteristic of different sources of uncertainty and, as such, informative of when and how to clarify, abstain, or answer.


\subsection{Constructing the Belief State}
An LLM with parameters $\theta$ defines a distribution $P_\theta(Y| X=x)$ over responses $Y\in \mathcal{X}$ given input $X=x \in \mathcal{X}$, where $\mathcal{X}$ is the space of strings. We take the stance that, for any input $X=x$, the probability measure $P_\theta$ under a given sampling algorithm is a representation of the LLM's uncertainty about the response $Y$, specifying the space of responses plausible under the model \cite{baan2023uncertainty}. 

To summarise the predictive distribution for a given input $X=x$ one can draw multiple samples using Monte-Carlo (MC) simulation. Practitioners often reduce this set of strings to a single number to power a decision making tool, like what generation to pick \cite{eikema-aziz-2020-map,wu2025better} or when to abstain \cite{farquhar2024detecting,vashurin-etal-2025-benchmarking}.

With BAG we retain the full set of $K$ samples as the \textit{belief state} $B_\theta(x) := (Y^{(1)}, \ldots, Y^{(K)})$, where $Y^{(k)} \sim P_\theta$. 
Interestingly, this representation not only captures diversity across generations that a model deems plausible, but also linguistic cues within generations, like hedging, refusals, clarifications, knowledge cut-offs, or nuanced acknowledgments of ambiguity. 

\subsection{Prompting for Conversational Strategies}
\label{sec:method_prompt}
BAG is a two-stage approach. First a model is shown a question to obtain multiple generations. Then these generations are included in a new prompt to help decide on and execute a conversational strategy. This prompt includes the user question $x$ and belief state $B$ and defines the allowed strategies, output format and how to analyse the belief state to arrive at a strategy. Appendix Figure \ref{fig:prompt-template}d shows an example prompt.

A powerful property of BAG is the ability to capture multiple intuitions behind popular uncertainty quantification methods in plain text, directly in the prompt. We explore four intuitions: (1) inconsistent facts among generations point to hallucinations and indicate a form of epistemic uncertainty that should lead to abstaining \cite[akin to semantic entropy;][]{kuhn2023semantic}, (2) generations that address different interpretations of the question should lead to a clarification question, (3) generations that are largely consistent should lead to a direct answer, and (4), direct answers should resemble the entire belief state \cite[akin to MBR;][]{eikema-aziz-2020-map}. We discuss the exact prompts in Section \ref{sec:bag_instantiation}.

\subsection{BAG+}
If there are multiple conversational turns, like when the assistant model asks a clarification question, one could apply BAG to every assistant turn, repeatedly augmenting the generation process---sampling $K$ new generations given the conversation history with another range of available strategies. We call this version BAG+. BAG+ is particularly useful when, even after a clarification interaction with the user, the assistant model is (too) uncertain to provide an answer. Appendix Figure \ref{fig:bag_overview} shows an overview of the entire pipeline. 

%% file: sections/4-experimental_setup.tex
\section{Experimental Setup}

\subsection{Dataset} 
We evaluate on real-world Google search questions from AmbigQA \cite{min-etal-2020-ambigqa}, for four reasons. First, it is the standard in ambiguity and clarification research \cite[e.g.,][]{gao-etal-2021-answering, min-etal-2021-joint, cole-etal-2023-selectively, kim-etal-2024-aligning, zhang2025modeling, zhang-choi-2025-clarify, berant2026learning}. Second, many popular alternatives (\eg CLAMBER, ASQA, CAmbigNQ, CondAmbigQA; \citealp{zhang-etal-2024-clamber,stelmakh-etal-2022-asqa,lee-etal-2023-asking,gao-etal-2023-enabling,li-etal-2025-condambigqa}) are derived from AmbigQA or its source Natural Questions \cite{kwiatkowski-etal-2019-natural}, for instance, SituatedQA \cite{zhang-choi-2021-situatedqa}, inheriting the same problems while adding little independent evidence. Third, AmbigQA contains uncertainty arising from the input due to ambiguity and underspecification, but also the model, due to the closed-book QA task that forces models to retrieve facts from their parameters. A dataset exhibiting only one does not allow evaluation of both clarify and abstain. Fourth, AmbigQA supplies unique annotations that we exploit as \textbf{user intents} to evaluate BAG rigorously and interactively (Section \ref{sec:evaluation}).


\paragraph{User intents.} AmbigQA annotates each question with all possible answers and provides multiple disambiguated questions for them. For instance, \textit{``Who is the female singer on Gimme Shelter?''} yields two disambiguated pairs: \textit{Who was the female singer on the recorded version of Gimme Shelter?} $\to$ Merry Clayton and \textit{Who was the female singer on Gimme Shelter on tour?} $\to$ Lisa Fischer. Following recent work, we treat each disambiguation-answer pair as a separate \textit{user intent} \cite{zhang2025modeling, zhang-choi-2025-clarify, berant2026learning}.
We evaluate on the full development set (the test set is not public), filtering out 170 questions with conflicting annotations, resulting in 1,832 questions. Roughly half (58.5\%) of the questions are ambiguous, and those have three intents on average. See Appendix~\ref{app:dataset_statistics} for more dataset statistics.

\subsection{Instantiating BAG}
\label{sec:bag_instantiation}
\paragraph{Conversational Strategies.} We experiment with three strategies for BAG: \textbf{clarify} to address uncertainty in the data (\eg ambiguity in the input), \textbf{abstain} to address uncertainty in the model (\eg about facts), and \textbf{direct answer} for little to no uncertainty. We leave other strategies, like retrieving external information (using BAG to decide when to RAG) or long-form answers, to future work. When applying BAG+ to the final assistant turn, we allow only the \textbf{direct answer} and \textbf{abstain} strategies, limiting the total number of conversational turns to four.

\paragraph{BAG prompts.} We test three prompts with slightly different instructions for how to analyse the belief state (BAG1-3). BAG1 in Appendix Figure \ref{fig:prompt-template}d shows the most basic variant, focusing on sample consistency. BAG2 in Appendix Figure \ref{fig:prompt-belief6} includes a Minimum Bayes Risk-like intuition, and further delineates when to abstain versus clarify. BAG3 in Appendix Figure \ref{fig:prompt-belief7} extends it with a two-step instruction: first cluster generations, then conjecture if multiple clusters represent different interpretations of the question or just factual inconsistencies (\ie hallucinations). We use a single prompt for BAG+, shown in Appendix Figure \ref{fig:prompt-final_bag}, and provide an overview of all prompts in Appendix Figure \ref{fig:bag_overview}.

\paragraph{Models.} We test models of various size and focus: the 7B and 13B OLMo2-instruct, 7B OLMo3-instruct, 8B and 14B Qwen3, and Gemini-2.5-Flash.\footnote{We disable thinking mode because of the extremely large token cost, especially in Olmo3 and Qwen3.}
We use $K=10$ unbiased samples to construct the belief state and use each model's recommended decoding setting (temperature, top-$k$, top-$p$ and min-$p$) to generate user-facing responses.


%% file: sections/5-evaluation.tex
\input{figures/combined_table}

\section{Evaluation}
\label{sec:evaluation}
We introduce four generation settings. For an overview of the prompts, see Appendix Figure \ref{fig:prompt-template}.

\paragraph{Direct generation.} The \textbf{standard baseline} generates a response given the original question. The \textbf{disambiguation oracle} uses an (annotated) disambiguated question from AmbigQA, if one exists, or the original question otherwise (collapsing to the standard baseline). We consider this an oracle for when and how to clarify, removing ambiguity and evaluating if models have the relevant factual knowledge to answer a particular intent.

\paragraph{Augmentation methods.} The main augmentation method  conditions on the belief state: \textbf{Belief-Augmented Generation (BAG)}, with an optional second round for turn 4:  \textbf{BAG+}. We also introduce \textbf{Strategy-Augmented Generation (SAG)}, a prompt-only baseline that receives the question and strategy instructions, but no belief state. \textbf{SAG+} is the application of this prompt-only baseline to the fourth turn, allowing for abstention after clarifications just like BAG+.

\paragraph{Simulating Interaction.}
When the assistant model asks a clarification question, we use Gemini-2.5-Flash to role-play a user. We prompt it with a user intent (hidden from the assistant model): a disambiguated version of the question if the annotation exists, and the reference answer if the question is not ambiguous. We design a conservative prompt (Figure \ref{fig:user-prompt}) that elicits ``I don't know'' if the clarification question is unanswerable or off topic to avoid rewarding bad clarification questions. Appendix \ref{app:user_error} provides an error analysis---we generally find high-quality user answers. 

We limit each conversation to four turns: user question, potential assistant clarification question, user response, assistant response. See Figure \ref{fig1:bag} for an example. We evaluate direct and clarify strategies against one specific user intent. Following standard selective prediction practices (\eg \citealp{kuhn2023semantic}), we exclude questions that a model abstains on and compute accuracy on the remaining subset. An increase in accuracy means that models abstain on questions that they would have gotten wrong otherwise---which is exactly the goal.

\subsection{Metrics}
\label{metrics}
We measure QA accuracy with a \textbf{human reference-based} LLM judge because 1) we found many false positives and false negatives with lexical metrics, 2) to mitigate hallucinations by the LLM judge. We employ Gemini-2.5-Flash. See Appendix Figure \ref{fig:judge-prompt} for the prompt. A small manual analysis detailed in Appendix \ref{app:judge_error_analysis} reveals a 97\% correctness rate of our reference-based LLM judge. In addition, we encourage the reader to manually explore questions, generations, user simulations, and judge verdicts with the \href{https://jorisbaan.nl/belief-augmented-generation}{BAG explorer \faDesktop}.

For the direct generation settings, we evaluate ambiguous questions with multiple intents against \textbf{one} (randomly chosen) intent and \textbf{any} of the possible intents.

\paragraph{One intent.} This is the core metric evaluating whether models give the correct answer to a specific user intent. This metric is strict and may be unfair without the potential for interaction, \ie it may reject generations that answer another intent, even though a user might judge the generation's relevance themselves, potentially following up by making their intent more explicit. 

\paragraph{Any intent.} Do models give a correct answer w.r.t. any of the possible intents? We consider this an upper bound that does not care about ambiguity.

%% file: figures/combined_table.tex
\begin{table*}[ht]
\centering\footnotesize
\setlength{\tabcolsep}{6pt}
\begin{tabular}{lccccccccc}
\toprule
 & \multicolumn{4}{c}{\textbf{Direct Generation}} & & \multicolumn{4}{c}{\textbf{Augmented Generation}} \\
\cmidrule(lr){2-5} \cmidrule(lr){7-10}
 & \multicolumn{2}{c}{One\,intent} & \multicolumn{2}{c}{Any\,intent} & & \multicolumn{4}{c}{One\,intent} \\
\cmidrule(lr){2-3} \cmidrule(lr){4-5} \cmidrule(lr){7-10}
 & Standard & Disambig & Standard & Disambig & & SAG+ & BAG1+ & BAG2+ & BAG3+ \\
\midrule
OLMo2-7B & \cellcolor[rgb]{0.647,0.000,0.149}\textcolor{white}{33.9} & \cellcolor[rgb]{0.964,0.477,0.286}{36.3} & \cellcolor[rgb]{0.000,0.408,0.216}\textcolor{white}{44.8} & \cellcolor[rgb]{0.710,0.876,0.454}{41.2} & & \cellcolor[rgb]{0.430,0.754,0.391}{41.5} & \cellcolor[rgb]{0.008,0.423,0.223}\textcolor{white}{44.7} & \cellcolor[rgb]{0.342,0.713,0.374}{42.8} & \cellcolor[rgb]{0.607,0.832,0.411}{41.7} \\
OLMo2-13B & \cellcolor[rgb]{0.647,0.000,0.149}\textcolor{white}{45.3} & \cellcolor[rgb]{0.993,0.702,0.397}{49.6} & \cellcolor[rgb]{0.000,0.408,0.216}\textcolor{white}{59.2} & \cellcolor[rgb]{0.489,0.780,0.398}{55.9} & & \cellcolor[rgb]{0.739,0.089,0.151}\textcolor{white}{46.8} & \cellcolor[rgb]{0.671,0.859,0.428}{54.9} & \cellcolor[rgb]{0.548,0.806,0.404}{55.6} & \cellcolor[rgb]{0.636,0.845,0.414}{55.1} \\
OLMo3-7B & \cellcolor[rgb]{0.647,0.000,0.149}\textcolor{white}{24.7} & \cellcolor[rgb]{0.678,0.030,0.150}\textcolor{white}{25.2} & \cellcolor[rgb]{0.982,0.607,0.346}{32.4} & \cellcolor[rgb]{0.939,0.390,0.246}{29.9} & & \cellcolor[rgb]{0.762,0.111,0.151}\textcolor{white}{29.5} & \cellcolor[rgb]{0.944,0.977,0.673}{39.8} & \cellcolor[rgb]{0.835,0.930,0.535}{41.8} & \cellcolor[rgb]{0.000,0.408,0.216}\textcolor{white}{52.9} \\
Qwen3-8B & \cellcolor[rgb]{0.647,0.000,0.149}\textcolor{white}{33.1} & \cellcolor[rgb]{0.686,0.037,0.150}\textcolor{white}{33.7} & \cellcolor[rgb]{0.995,0.825,0.500}{43.2} & \cellcolor[rgb]{0.980,0.597,0.341}{40.3} & & \cellcolor[rgb]{0.739,0.089,0.151}\textcolor{white}{34.6} & \cellcolor[rgb]{0.000,0.408,0.216}\textcolor{white}{60.2} & \cellcolor[rgb]{0.998,0.926,0.625}{45.0} & \cellcolor[rgb]{0.986,0.637,0.360}{40.8} \\
Qwen3-14B & \cellcolor[rgb]{0.647,0.000,0.149}\textcolor{white}{40.1} & \cellcolor[rgb]{0.859,0.221,0.168}\textcolor{white}{42.8} & \cellcolor[rgb]{0.974,0.989,0.713}{52.4} & \cellcolor[rgb]{0.996,0.855,0.526}{49.3} & & \cellcolor[rgb]{0.894,0.296,0.202}{42.8} & \cellcolor[rgb]{0.000,0.408,0.216}\textcolor{white}{63.8} & \cellcolor[rgb]{0.430,0.754,0.391}{58.8} & \cellcolor[rgb]{0.225,0.656,0.344}{60.4} \\
Gemini-2.5-Flash & \cellcolor[rgb]{0.647,0.000,0.149}\textcolor{white}{59.1} & \cellcolor[rgb]{0.880,0.950,0.585}{68.7} & \cellcolor[rgb]{0.000,0.408,0.216}\textcolor{white}{75.6} & \cellcolor[rgb]{0.084,0.563,0.296}{74.2} & & \cellcolor[rgb]{0.999,0.983,0.721}{67.4} & \cellcolor[rgb]{0.869,0.945,0.569}{68.8} & \cellcolor[rgb]{0.749,0.893,0.479}{69.8} & \cellcolor[rgb]{0.904,0.959,0.617}{68.4} \\
\bottomrule
\end{tabular}
\caption{Accuracy on AmbigQA for direct generation (left) and augmented generation (right) in both the first and final assistant turns (SAG+/BAG+). All BAG+ prompt/model configurations outperform standard generation and the prompt-only baseline SAG+. For 17/18 configurations, BAG+ also outperforms the disambiguation oracle. Colours are row-normalised.}
\label{tab:bag_results}
\end{table*}

%% file: sections/6-results-direct.tex
\section{Results}
\label{sec:direct_results}

We first analyse direct generation results, with no possibility to clarify or abstain. See Table~\ref{tab:bag_results}, left. We care about the relative differences between generation settings, not models. Therefore, colours are row-normalised.

\paragraph{Disambiguations cannot fix lack of knowledge.} 
The disambiguation oracle yields relatively small improvements over the standard single intent baseline for some models (\eg Qwen3-8B), and much larger improvements for others (\eg Gemini-2.5-Flash). This demonstrates that some models are unable to exploit disambiguations due to a lack of factual knowledge, rather than an inability to handle ambiguity. This indicates varying and sometimes limited potential for clarification, further motivating why BAG supports both clarification and abstention.

\paragraph{Specific intents are difficult.} 
The gap between the standard one vs.~any intent metric reflects the inherent difficulty of recovering a specific user intent without interaction or disambiguation---a model may give a perfectly reasonable answer while missing the target interpretation. Still, the one intent metric is valuable when considering that users may ask an ambiguous question while having a specific intent in mind, especially if they cannot foresee which interpretation is salient to the model.

\paragraph{Models presuppose a ``salient'' interpretation.}
The disambiguation oracle's improvement from the one intent to any intent metric indicates that models may presuppose a salient interpretation so strongly that they ignore the disambiguation, perhaps not having the relevant factual knowledge to support other intents. This aligns with \citet{sedova-etal-2024-know} who find biases toward preferred readings of entities that correlate with the frequency of those readings in training data. This may also limit the effect that clarifications have on changing such salient interpretations.\footnote{To investigate to what extent interpretations salient to models align with interpretations salient to humans, we need more and more diverse annotations than AmbigQA offers.}

\paragraph{Performance differences across models.} 
Finally, while not relevant to evaluating BAG, Table~\ref{tab:bag_results} shows that Gemini-2.5-Flash performs best with 59.1\% accuracy on the one intent metric, followed by Olmo2. Olmo2 outperforms Qwen3 and Olmo3, which is consistent with results from the original authors on PopQA \citep{olmo3-2025} and may reflect the newer Qwen3's and Olmo3's relative focus on reasoning and instruction following over factual knowledge. Gemini-2.5-flash is also better at exploiting disambiguations, with an accuracy increase of 9.6 points. Since we do not have access to training data, there are few conclusions we can draw from this. Moreover, while it may seem that the Gemini-2.5-Flash judge is unfairly favouring itself, recall that we use it as a reference-based LLM judge, with explicit instructions to use the AmbigQA references as authoritative (Section \ref{metrics}). 

%% file: sections/7-results-bag.tex
\begin{figure}[t]
    \centering
    \includegraphics[width=0.99\linewidth]{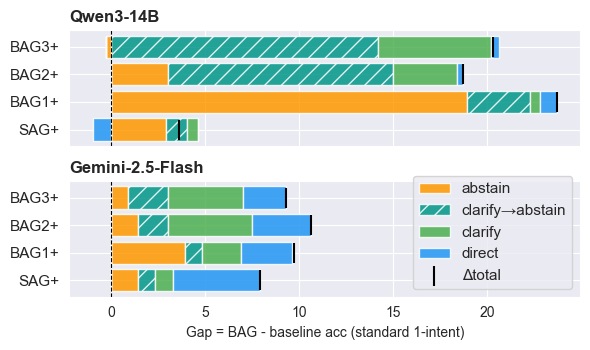}
    \caption{The contribution of each strategy to BAG+'s total accuracy increase (shown on the x-axis). Abstain is the driving force for Qwen3, while for Gemini the contributions are more balanced. }
    \label{fig:bag_decomposed}
\end{figure}

\subsection{Augmented Generation}
\label{sec:bag_results}
Table \ref{tab:bag_results} (right) reports our main results for belief-augmentation of both the first and the final assistant turn (BAG+). To isolate the contribution of the first round, Appendix Table \ref{tab:bagnonplus_results} reports BAG applied to the first assistant turn only.

A single round of BAG yields substantial gains: it outperforms the standard one intent baseline in 17 out of 18 model–prompt configurations (the exception being Qwen3-8B with BAG3); the prompt-only SAG in 15 out of 18; and even the disambiguation oracle in 11 out of 18. The best prompt variant per model improves over the standard one intent baseline by 12.4 points averaged across models (a 34\% relative improvement), ranging from 5.8 (OLMo2-7B) to 25.7 (Qwen3-8B), and over SAG by 10.2 points.

Augmenting the final assistant turn as well (BAG+) improves results further: in Table \ref{tab:bag_results} (right), all BAG1–3+ variants outperform both the standard baseline and SAG+; and BAG+ outperforms the disambiguation oracle in 17 out of 18 configurations. The best prompt variant per model improves over the standard baseline by 18.5 points averaged across models (a 55\% relative improvement), and over the disambiguation oracle by 15.1 points. We attribute this to the assistant model often lacking the relevant factual knowledge to answer correctly even after clarification---something we also observed for the disambiguation oracle in Section \ref{sec:direct_results}. BAG+ is able to catch this in its updated belief state and abstains rather than incorrectly guessing.

\subsubsection{Decomposing BAG Performance}
\begin{figure*}
    \centering
    \includegraphics[width=0.99\linewidth]{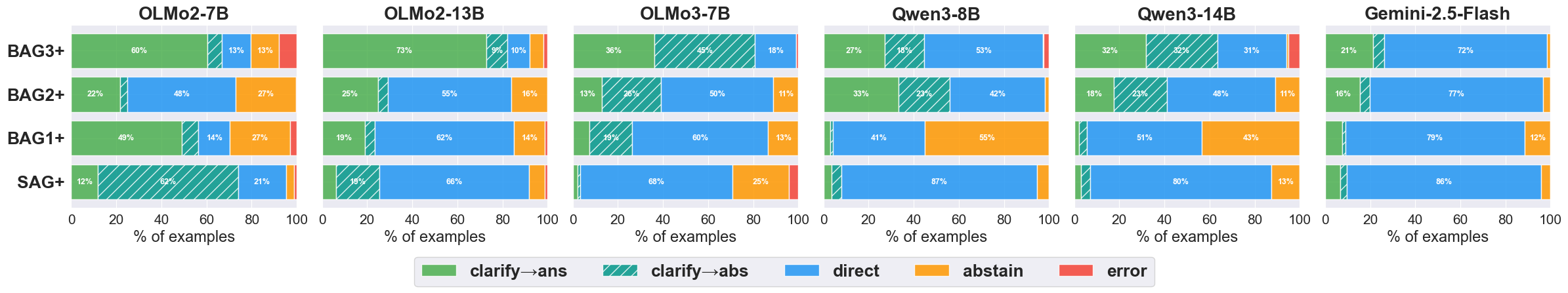}
    \caption{The strategies that models pick can vary a lot based on the model class and instructions. For example, larger models often choose to answer directly across prompts, indicating less uncertainty, whereas OLMo2-7B clarifies a lot.}
    \label{fig:strategies}
\end{figure*}

To understand if and how each strategy contributes, we decompose BAG+'s accuracy improvement over the standard direct generation baseline in Figure \ref{fig:bag_decomposed}. Orange shows the contribution of abstaining (by excluding those examples from evaluation); green the improvement of the final answer after clarification over the baseline on the clarify-subset; and blue of the direct answers relative to the standard baseline on the direct-subset. Every strategy contributes positively but to varying degree. For Qwen, abstaining is the driving force behind its accuracy improvement, in line with its low baseline and disambiguation oracle scores in Table \ref{fig1:bag}. Appendix Figures \ref{fig:decomp_all_models},\ref{fig:decomp_all_nonplus} show other models.

\subsubsection{Strategy routing}
This decomposition captures both \textit{what} strategy is chosen and the \textit{quality} of the actual response. We next isolate \textit{strategy routing}: how often and when are strategies chosen? Figure \ref{fig:strategies} shows the frequency of each strategy per model and prompt. While the direct answer strategy is generally most often chosen there are big differences across models. We hypothesise that this is due to structurally different belief states with varying levels of consistency (\ie uncertainty) and verbosity, which we further explore in Section \ref{sec:faithfulness}. For example, the best model, Gemini-2.5-Flash, most often answers directly, exhibiting less uncertainty.

There are also differences across prompts. These are partly by design: \eg BAG3 instructs models to cluster generations and conjecture if and which interpretation of the user's question might lead to each cluster. Models can almost always come up with such interpretations, which explains the high rate of clarification. However, instruction following is also a factor: Gemini-2.5-Flash---the best instruction follower---shows substantially less variation across BAG prompt variants. Next, we attempt to evaluate the strategy routing: on what kind of questions was each strategy chosen?

\paragraph{Accuracy.} Table \ref{tab:routing_profile} shows accuracy of the non-augmented, standard direct generation baseline on strategy subsets defined by augmented generation in the first assistant turn. This tells us if the harder questions are rightfully being routed to clarify or abstain, while the easier ones are answered directly. Indeed, baseline accuracy on the direct answer subset \textit{acc(D)} is much higher than on the abstain subset, sometimes as much as 5-7x (Qwen3), whereas the clarified-subset accuracy lies in between. This indicates that augmented generation leads to sensible strategy decisions. Also, belief-augmented prompts make better routing decisions than the prompt-only SAG variant, demonstrated by higher accuracy on the direct and lower accuracy on the abstain subset.

\paragraph{Ambiguity.} The clarify subsets contain the highest share of ambiguous questions across models, shown by the \textit{amb(C)} column being dark blue. However, this share is still only 53.5-76.7\%, which is consistent with related work showing that detecting ambiguity or when to clarify is a very challenging task \cite{zhang-etal-2024-clamber,tanjim-etal-2025-disambiguation, zhang2025modeling}. We posit that ambiguity detection in factual QA is inherently very difficult: recognizing multiple intents given a deceptively straightforward question requires both an understanding of ambiguity and highly specialised factual knowledge (\eg that a different person performed a song on tour), and we suspect that even humans would struggle to mark these questions as ambiguous. That said, since LLMs are increasingly used as a substitute for search engines~\cite{NBERw34255}, the bar may arguably be higher.

\input{figures/routing_table}

\subsection{Effect of Clarification} 
The ideal conditions to clarify are:
\begin{enumerate*}[label=(\arabic*)]
    \item a direct answer would have been wrong,
    \item the clarification question is relevant and a user would be able to answer it,
    \item without the user's answer, a model could not give the correct answer (\eg RAG would not help since a model does not know what information to fetch or fetches conflicting information), and finally
    \item the model has the factual knowledge to answer the disambiguated question.
\end{enumerate*}

Section \ref{sec:direct_results} shows that condition 1 is often met (direct generation accuracy is low) and condition 4 is not (factual knowledge, not ambiguity, is often the problem, for which abstain is a better strategy). This limits the potential for clarification, which we confirm in Figure \ref{fig:bag_decomposed}. We will qualitatively investigate conditions 2,3 next. 

Belief states display both \textit{interpretation variation} due to multiple plausible interpretations of the question, and \textit{hallucination variation} due to inconsistent facts, falling neatly on opposite ends of the data-model uncertainty spectrum \cite{baan2023uncertainty}. Interpretation variation should be addressed by interacting with the user, and hallucination variation by abstaining. However, we find that detecting this difference is challenging. 
BAG guides models to formulate a clarification question based on their belief state. When the belief state contains multiple plausible answers, each contextualised (\eg \textit{"in the \underline{US}, ..."} vs \textit{"in \underline{Canada}, ..."}, or \textit{"the \underline{FIFA} world cup ..."} vs \textit{"The \underline{rugby} world cup..."}), the clarifications are generally useful, like the examples in Figures~\ref{fig1:bag}-\ref{fig:bag_3examples}.

However, models sometimes come up with non-existent ambiguity to explain hallucination variation, especially if generations are not meaningfully contextualised. For example, non-existing adaptations of a movie in which different actors play the same character to explain mentions of those actors in the belief state. Put differently: clarification questions reflect what \citet{kim-etal-2024-aligning} call \textit{perceived ambiguity}, which does not necessarily align with how humans perceive ambiguity in the real world. While such clarification questions are truthful reflections of what models deem plausible---and arguably a better strategy than responding with one of the incorrect generations---they may still confuse a user.  



\subsection{Computational Cost}
BAG's computational cost, like other sample-based approaches (\eg MBR, semantic entropy, self-consistency, verbalised uncertainty) is $K$ samples, plus an additional prompt to reason over them. The optional BAG+ adds another round. However, BAG goes beyond a single decision by simultaneously determining 1) when to answer (selective prediction), 2) how to answer (self-consistency, MBR, verbalised uncertainty), and 3) when and how to clarify. 

Preliminary tests reveal that Qwen3 thinking mode consumes 7x more output tokens than direct generation, for just 2 more accuracy points. This cost is not much higher than sampling 10 generations, which, unlike thinking, is parallelisable. This puts BAG's costs in perspective.

To reduce token cost, we study the effect of two brevity-inducing prompts: \textit{"Please provide a concise answer to the following question:"}, and \textit{"(...) short answer of at most 1 sentence (...)"}. They reduce average generation length by up to a factor 10 without severely impacting accuracy (Appendix Table \ref{tab:bag_brevity_results}). However, the verbal cues for how models interpret a question (\eg \textit{"the \textit{FIFA} world cup was in ..."}) are often lost (examples in Appendix Table \ref{table:examples_1sentence}), reducing the means by which users can judge an answer, and BAG a belief state.


%% file: figures/routing_table.tex
\begin{table}[ht]
\centering\scriptsize
\setlength{\tabcolsep}{4pt}
\begin{tabular}{ll|cccccc}
\toprule
 &  & acc(C) & acc(D) & acc(A) & amb(C) & amb(D) & amb(A) \\
\midrule
\multirow{2}{*}{\shortstack[l]{OL2\\7B}} & S & \cellcolor[rgb]{0.709,0.059,0.150}\textcolor{white}{32.6} & \cellcolor[rgb]{0.000,0.408,0.216}\textcolor{white}{38.6} & \cellcolor[rgb]{0.647,0.000,0.149}\textcolor{white}{32.4} & \cellcolor[rgb]{0.969,0.984,1.000}\textcolor{black}{53.5} & \cellcolor[rgb]{0.623,0.793,0.883}\textcolor{black}{57.1} & \cellcolor[rgb]{0.031,0.188,0.420}\textcolor{white}{63.2} \\
\cline{2-8}
 & B & \cellcolor[rgb]{0.647,0.000,0.149}\textcolor{white}{28.7} & \cellcolor[rgb]{0.000,0.408,0.216}\textcolor{white}{45.8} & \cellcolor[rgb]{0.912,0.334,0.220}\textcolor{black}{31.5} & \cellcolor[rgb]{0.031,0.188,0.420}\textcolor{white}{55.1} & \cellcolor[rgb]{0.969,0.984,1.000}\textcolor{black}{52.5} & \cellcolor[rgb]{0.147,0.460,0.719}\textcolor{black}{54.4} \\
\arrayrulecolor{black!35}\specialrule{1pt}{2pt}{2pt}\arrayrulecolor{black}
\multirow{2}{*}{\shortstack[l]{OL2\\13B}} & S & \cellcolor[rgb]{0.921,0.352,0.228}\textcolor{black}{35.5} & \cellcolor[rgb]{0.000,0.408,0.216}\textcolor{white}{50.0} & \cellcolor[rgb]{0.647,0.000,0.149}\textcolor{white}{32.5} & \cellcolor[rgb]{0.316,0.612,0.799}\textcolor{black}{53.8} & \cellcolor[rgb]{0.031,0.188,0.420}\textcolor{white}{55.4} & \cellcolor[rgb]{0.969,0.984,1.000}\textcolor{black}{51.6} \\
\cline{2-8}
 & B & \cellcolor[rgb]{0.990,0.667,0.373}\textcolor{black}{37.5} & \cellcolor[rgb]{0.000,0.408,0.216}\textcolor{white}{57.0} & \cellcolor[rgb]{0.647,0.000,0.149}\textcolor{white}{29.4} & \cellcolor[rgb]{0.031,0.188,0.420}\textcolor{white}{56.4} & \cellcolor[rgb]{0.969,0.984,1.000}\textcolor{black}{51.5} & \cellcolor[rgb]{0.657,0.808,0.895}\textcolor{black}{53.2} \\
\arrayrulecolor{black!35}\specialrule{1pt}{2pt}{2pt}\arrayrulecolor{black}
\multirow{2}{*}{\shortstack[l]{OL3\\7B}} & S & \cellcolor[rgb]{0.693,0.044,0.150}\textcolor{white}{19.4} & \cellcolor[rgb]{0.000,0.408,0.216}\textcolor{white}{27.0} & \cellcolor[rgb]{0.647,0.000,0.149}\textcolor{white}{19.2} & \cellcolor[rgb]{0.031,0.188,0.420}\textcolor{white}{58.1} & \cellcolor[rgb]{0.969,0.984,1.000}\textcolor{black}{54.2} & \cellcolor[rgb]{0.216,0.529,0.754}\textcolor{black}{56.8} \\
\cline{2-8}
 & B & \cellcolor[rgb]{0.850,0.202,0.159}\textcolor{white}{13.9} & \cellcolor[rgb]{0.000,0.408,0.216}\textcolor{white}{36.9} & \cellcolor[rgb]{0.647,0.000,0.149}\textcolor{white}{11.1} & \cellcolor[rgb]{0.031,0.188,0.420}\textcolor{white}{55.9} & \cellcolor[rgb]{0.106,0.413,0.686}\textcolor{white}{54.3} & \cellcolor[rgb]{0.969,0.984,1.000}\textcolor{black}{48.7} \\
\arrayrulecolor{black!35}\specialrule{1pt}{2pt}{2pt}\arrayrulecolor{black}
\multirow{2}{*}{\shortstack[l]{Qw3\\8B}} & S & \cellcolor[rgb]{0.770,0.118,0.151}\textcolor{white}{14.5} & \cellcolor[rgb]{0.000,0.408,0.216}\textcolor{white}{36.1} & \cellcolor[rgb]{0.647,0.000,0.149}\textcolor{white}{13.0} & \cellcolor[rgb]{0.031,0.188,0.420}\textcolor{white}{64.1} & \cellcolor[rgb]{0.435,0.691,0.843}\textcolor{black}{54.4} & \cellcolor[rgb]{0.969,0.984,1.000}\textcolor{black}{45.0} \\
\cline{2-8}
 & B & \cellcolor[rgb]{0.832,0.177,0.153}\textcolor{white}{13.2} & \cellcolor[rgb]{0.000,0.408,0.216}\textcolor{white}{57.2} & \cellcolor[rgb]{0.647,0.000,0.149}\textcolor{white}{8.6} & \cellcolor[rgb]{0.031,0.188,0.420}\textcolor{white}{57.0} & \cellcolor[rgb]{0.179,0.493,0.735}\textcolor{black}{55.2} & \cellcolor[rgb]{0.969,0.984,1.000}\textcolor{black}{51.0} \\
\arrayrulecolor{black!35}\specialrule{1pt}{2pt}{2pt}\arrayrulecolor{black}
\multirow{2}{*}{\shortstack[l]{Qw3\\14B}} & S & \cellcolor[rgb]{0.647,0.000,0.149}\textcolor{white}{18.6} & \cellcolor[rgb]{0.000,0.408,0.216}\textcolor{white}{45.2} & \cellcolor[rgb]{0.716,0.066,0.150}\textcolor{white}{19.6} & \cellcolor[rgb]{0.031,0.188,0.420}\textcolor{white}{69.0} & \cellcolor[rgb]{0.799,0.874,0.945}\textcolor{black}{54.2} & \cellcolor[rgb]{0.969,0.984,1.000}\textcolor{black}{50.0} \\
\cline{2-8}
 & B & \cellcolor[rgb]{0.793,0.140,0.152}\textcolor{white}{16.5} & \cellcolor[rgb]{0.000,0.408,0.216}\textcolor{white}{68.5} & \cellcolor[rgb]{0.647,0.000,0.149}\textcolor{white}{12.2} & \cellcolor[rgb]{0.031,0.188,0.420}\textcolor{white}{61.3} & \cellcolor[rgb]{0.969,0.984,1.000}\textcolor{black}{53.7} & \cellcolor[rgb]{0.959,0.978,0.997}\textcolor{black}{53.8} \\
\arrayrulecolor{black!35}\specialrule{1pt}{2pt}{2pt}\arrayrulecolor{black}
\multirow{2}{*}{\shortstack[l]{Gem\\2.5Flash}} & S & \cellcolor[rgb]{0.986,0.637,0.360}\textcolor{black}{37.5} & \cellcolor[rgb]{0.000,0.408,0.216}\textcolor{white}{63.1} & \cellcolor[rgb]{0.647,0.000,0.149}\textcolor{white}{27.3} & \cellcolor[rgb]{0.031,0.188,0.420}\textcolor{white}{76.7} & \cellcolor[rgb]{0.760,0.852,0.932}\textcolor{black}{52.8} & \cellcolor[rgb]{0.969,0.984,1.000}\textcolor{black}{44.2} \\
\cline{2-8}
 & B & \cellcolor[rgb]{0.989,0.657,0.369}\textcolor{black}{33.5} & \cellcolor[rgb]{0.000,0.408,0.216}\textcolor{white}{67.3} & \cellcolor[rgb]{0.647,0.000,0.149}\textcolor{white}{19.5} & \cellcolor[rgb]{0.031,0.188,0.420}\textcolor{white}{63.9} & \cellcolor[rgb]{0.498,0.725,0.856}\textcolor{black}{53.6} & \cellcolor[rgb]{0.969,0.984,1.000}\textcolor{black}{45.1} \\
\bottomrule
\end{tabular}
\caption{How good is the choice of strategy in the first assistant turn? We group questions by the strategy a model predicts (\underline{\textbf{C}}larify / \underline{\textbf{D}}irect / \underline{\textbf{A}}bstain) and report standard baseline \underline{\textbf{acc}}uracy (left) and share of \underline{\textbf{amb}}iguous questions (right) per group. For example, \textbf{acc(C)} is standard baseline accuracy on questions the model decides to clarify. Ideally \textbf{acc(D)} is high, \textbf{acc(A)} and \textbf{acc(C)} low, and \textbf{amb(C)} high. Colours are row-normalised. S: \textbf{S}AG, B: average over \textbf{B}AG1--3.}
\label{tab:routing_profile}
\end{table}

%% file: sections/8-faithfulness.tex
\section{Faithfulness}
\label{sec:faithfulness}

We evaluated BAG in terms of QA performance, but equally important is: are its strategic decisions faithful to its internal distribution?

We cluster generations on the main answer they assert and compute semantic entropy \cite{kuhn2023semantic} with Gemini-2.5-Flash. We find that belief states range from just a single answer to ten totally distinct answers. For smaller models, belief states are almost uniformly distributed across this range, while Gemini-2.5-Flash has many more single-answer belief states; about 50\%; see Appendix Figure \ref{fig:entropy}.

\begin{figure}
    \centering
    \includegraphics[width=0.99\linewidth]{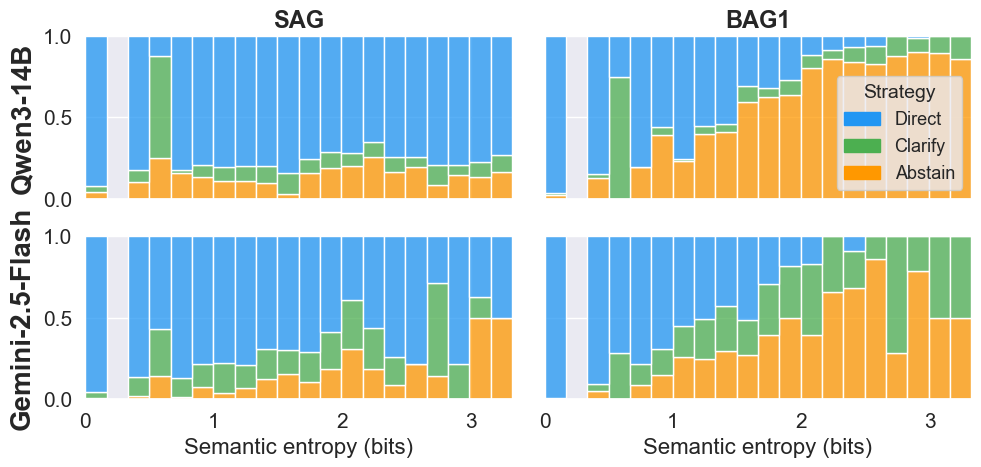}
    \caption{Routing decisions vs belief state entropy: BAG is more faithful to the predictive distribution than SAG, answering directly when entropy is low, increasingly clarifying and abstaining as entropy increases.}
    \label{fig:h_claim}
\end{figure}

\paragraph{BAG's strategies are more faithful.} 
Our hypothesis is that models will choose the direct answer strategy for lower entropy, and clarify or abstain for higher entropy belief states. Indeed, Figure \ref{fig:h_claim} confirms this for Qwen3-14B and Gemini-2.5-Flash with BAG. Interestingly, the prompt-only SAG is not aligned with belief state entropy, providing further evidence that prompt-based uncertainty methods do not exploit the `internal' predictive distribution. We observe similar patterns for Olmo3 and Qwen3-8B, but not for Olmo2, which we explain by its limited ability to follow instructions.

\paragraph{Baselines rarely clarify or abstain.}
Direct generation without any augmentation strategy rarely clarifies or abstains at all: classifying generations with Gemini-2.5-Flash we find they only do so on 3\% of the questions. A small manual analysis (Appendix Figure \ref{fig:strategy_annotation}) confirms this, aligning with findings from previous work \cite{deng-etal-2023-prompting, shaikh-etal-2024-grounding, zhang-etal-2024-clamber, testoni-etal-2025-racquet}. This confirms that direct generation is even less faithful to the internal distribution.

\paragraph{Clarification reduces entropy.} 
We sample $K$ responses after clarification interactions (similar to BAG+), cluster them, and compute the difference between their semantic answer entropy and entropy of the turn 2 responses reported earlier. We find that clarification reduces entropy for all models. Averaged across BAG1-3, Gemini has the largest reduction of 54\%, followed by Olmo2-13B at 20\%, Qwen3-14B at 18\%, Qwen3-8B at 16\%, Olmo2-7B at 14\%, and Olmo3-7B at 1.7\%. This aligns with the disambiguation oracle results, which indicated that some models do not have the factual knowledge to exploit even annotated disambiguations.

%% file: sections/10-conclusion.tex
\section{Conclusion}
Belief-augmented generation unifies disparate sampling-based uncertainty methods into a single decision making framework. It combines a probabilistic view with modern instruction and reasoning capabilities to decide \textit{when} and \textit{how} to answer directly, clarify, or abstain. In our interactive ambiguous QA setting, BAG improves accuracy across six models and produces strategic decisions more faithful to the belief state than the prompt-only baselines, while standard direct generation rarely clarifies or abstains at all. Disentangling (semantic) variation in belief states due to hallucinations versus different interpretations---and thus when to clarify versus abstain---remains challenging. 



There are many promising avenues for future work. For example, experimenting with different strategies (\eg tool calling), deploying BAG in more autonomous, task-driven settings (\eg booking a trip with ambiguous instructions), and enriching the belief state (\eg including thinking traces, combining the power of reasoning models with a more traditional statistical perspective).

%% file: sections/11-acknowledgements.tex
\section{Acknowledgements}
We thank Dennis Ulmer and Evgenia Ilia for their valuable suggestions and insightful discussions, the members of Amsterdam's Dialogue Modelling Group and Mulini lab for their feedback, as well as the anonymous reviewers and meta-reviewer. This project received funding from the ELLIS Amsterdam Unit. BP is supported by the ERC Consolidator Grant DIALECT 101043235. 

%% file: sections/12-limitations.tex
\section{Limitations}
While AmbigQA is well suited to simulate different user intents due to its multiple disambiguation annotations, being similarly used by several recent studies due to a lack of good alternatives, its references can be outdated and its disambiguations slightly artificial: rather than the actual intent of the user who asked a question---impossible to reconstruct after the fact---AmbigQA instead collects all possible answers via Wikipedia search and retroactively create (minimally edited) disambiguated versions of the original question. An important avenue of research is therefore to collect new datasets with real world questions and intents that correspond to actual user intents, ideally from users with diverse cultural backgrounds (\eg AmbigQA is very US-centered), which also allows comparison between models' suppositions and whether they align with dominant user intents. 

Another limitation is the quality of the belief states, \ie the model's representation of uncertainty. While belief states provide richer insight into the range of responses probable under a model, and are successfully exploited for decision making tasks like decoding or selective prediction, they still merely reflect probability under a model, not reality (\eg highly consistent belief states with a single main claim can still be factually incorrect). It is important to improve the quality of belief states, which can then be more effectively exploited by methods such as BAG.

Finally, by using BAG as a test-time scaling method we incur a computational cost during inference. However, BAG can be extended for synthetic data creation, allowing models to be finetuned on uncertainty-guided strategies without additional inference costs.

%% file: sections/appendix.tex
\appendix

\input{figures/generation-settings.tex}
\input{figures/bag_prompts}

\section{Dataset Statistics}
\label{app:dataset_statistics}
Each AmbigQA question is annotated with a single answer (\ie intent), marking the question as unambiguous, or multiple disambiguated question-answer pairs (\ie, intents), marking it as ambiguous. The dataset contains 10,036 train and 2,002 development questions, with 10,251 and 3,312 annotations respectively (multiple annotators may annotate the same question). The development set has a higher ambiguity rate (58.5\% vs.\ 47.3\% for train) and annotation clash rate (\ie questions with both an ambiguous and unambiguous annotation: 8.5\% vs.\ 0.1\% for train---we filter those questions out). We evaluate on the full development set (the test set is not public), filtering out 170 questions with conflicting annotations, resulting in 1,832 questions. Questions average 9.0 words and answers 2.5 words across both splits. Ambiguous questions have on average 3.17 (±2.03) intents in dev and and 2.94 (±1.33) in train.

\section{LLM User Answer Simulator Error Analysis}
\label{app:user_error}
\input{figures/user_prompt}

The user answer simulation prompt for Gemini-2.5-flash is shown in Figure \ref{fig:user-prompt}. We find that simulated user answers are generally of high quality. The user simulator is explicitly instructed not to reveal the reference answer and only to answer the clarification question. Nevertheless, references occasionally leak through. On average, across all assistant models and prompts, the user answer contains the reference answer for 1.7\% of clarification questions. 

However, some clarification questions already contain the reference answer themselves, which is not a leak on the user simulator's part. Excluding those cases, only 0.7\% of user answers constitute true leaks. To assess their impact on our results, we count how often leaks co-occur with a positive clarification flip (direct answer incorrect → final answer correct), finding this in only 1.9\% of such flips. We therefore conclude that reference leaks do not meaningfully inflate the reported clarification gains.

The prompt instructs the user simulator not to answer clarification questions that are irrelevant to its intent or off-topic, to avoid rewarding bad CQs with useful additional information about the users intent. On average, about 5\% of the user responses are an exact match with "I don't know". This simulation strategy is very conservative, and we suspect real users to be much more helpful. 

For example, consider the AmbigQA question \textit{"What is the dog called in midsomer murders?"} with intent \textit{"What is the dog's breed called in midsomer murders?"} and the model's clarification question \textit{"There isn't a single dog that consistently appears throughout all of Midsomer Murders with one name, as there have been different dogs belonging to the various DCI Barnabys. To give you the most accurate answer, could you tell me which DCI Barnaby you're thinking of (Tom Barnaby, played by John Nettles, or John Barnaby, played by Neil Dudgeon)?"}. The user simulator answers \textit{"I don't know"}, with the following reasoning \textit{"Internal reasoning: The user is asking about the dog's breed, not its name. The AI's clarification question is about the DCI Barnaby to identify the dog's name. Since the user is asking about the breed, the specific DCI Barnaby is not relevant to the breed of dog. Therefore, the AI's clarification question is not helpful for the user's actual question."}. A real user might just respond with the much more helpful \textit{"Actually, I'm looking for the breed"}.

\section{LLM Judge Error Analysis}
\label{app:judge_error_analysis}

\input{figures/judge_prompt}
\begin{figure*}
    \centering
    \includegraphics[width=0.99\linewidth]{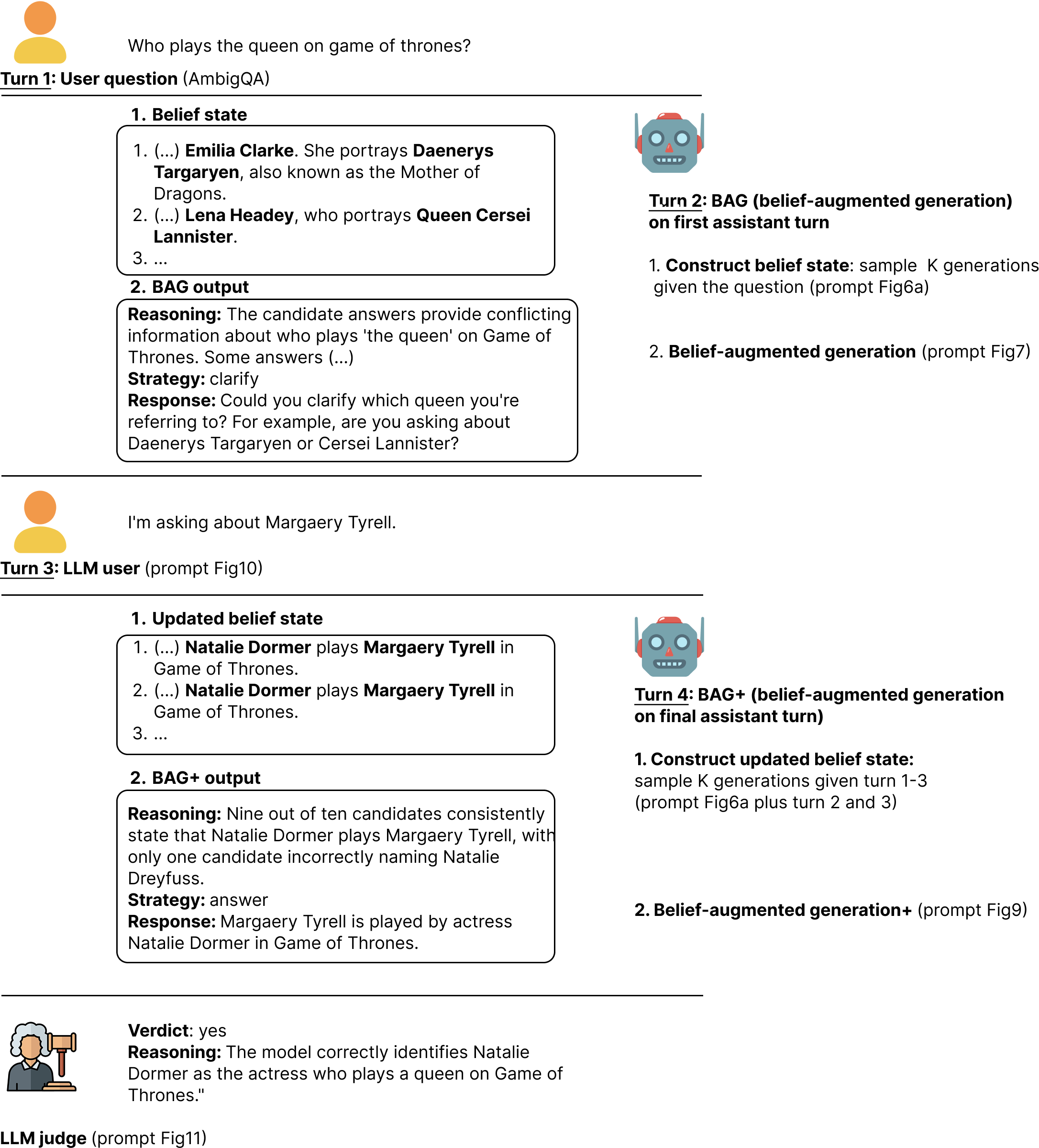}
    \caption{An overview of all BAG(+) pipeline steps, referencing each prompt used. Real AmbigQA dev example with ID 3312525434538504362. Generation output from Qwen3-14B with concise brevity induced generation, BAG2 (Fig. 7) and BAG+ (Fig. 9) prompts, and Gemini-2.5-Flash as user simulator (Fig. 10), and judge (Fig. 11).}
    \label{fig:bag_overview}
\end{figure*}

Figure \ref{fig:judge-prompt} shows the two prompts we use for the reference-based LLM judge. We generally find the judge verdicts of Gemini-2.5-Flash about whether a generation is correct with respect to a reference of high quality, and manually evaluate 30 direct generation answers from Olmo2-13B against the same single reference answer that the judge is evaluating against (the 1 intent setting, \ie leftmost prompt in Figure \ref{fig:judge-prompt}). We find just one error, \ie a success rate of 97\%. 

This error is actually an interesting case where the judge ignores the reference and uses its own knowledge to judge the generation as correct for the question "The cat in the hat boy and girl?". In fact, the generation does correctly answer another (much more obvious) interpretation of the question (the boy and the girl characters are unnamed), while being explicit about doing so, which makes it hard to fault the generation. This is a limitation of single-intent evaluation setting, because a real user might be able to judge that the generation is answering a different interpretation of the question than what they actually intended (\eg the annotated disambiguation "Who plays the girl in the 2003 Cat in the Hat film?"), and would potentially follow up to clarify this to the model.

\subsection{Reference Quality}
A more important limitation of the LLM judge is not its ability to judge whether a generation contains a reference, but the quality of the AmbigQA reference itself. In Table \ref{tab:ambigqa-examples} we characterise several types of questions and their annotated disambiguation-reference pairs. One notable category is questions with answers that change over time.

\paragraph{Outdated references.}
This problem is universal across QA datasets and discussed by \citet{pletenev-etal-2025-will}. We apply their pretrained classifier to filter out time-sensitive questions, and find it discards about 30\%. When we filter those out, QA accuracy increases with a few points across all models and configurations, but the relative differences between generation settings remain the same. We hypothesise that the effect is so small because all methods are impacted equally by the filtering. We choose not to include these results in the paper because we manually observed that many subtly non-mutable questions are also being filtered out (\eg "Who did Samuel L. Jackson play in Star Wars?", or "Who played James Bond in Licence to Kill?"). 

\paragraph{Artificial ambiguity.}
One issue with the dataset is that the disambiguation annotations reflect what annotators could find on Wikipedia rather than what interpretation users genuinely meant when they asked their question to Google Search. 


\input{figures/bag_results_nonplus}

\begin{figure}
    \centering
    \includegraphics[width=0.99\linewidth]{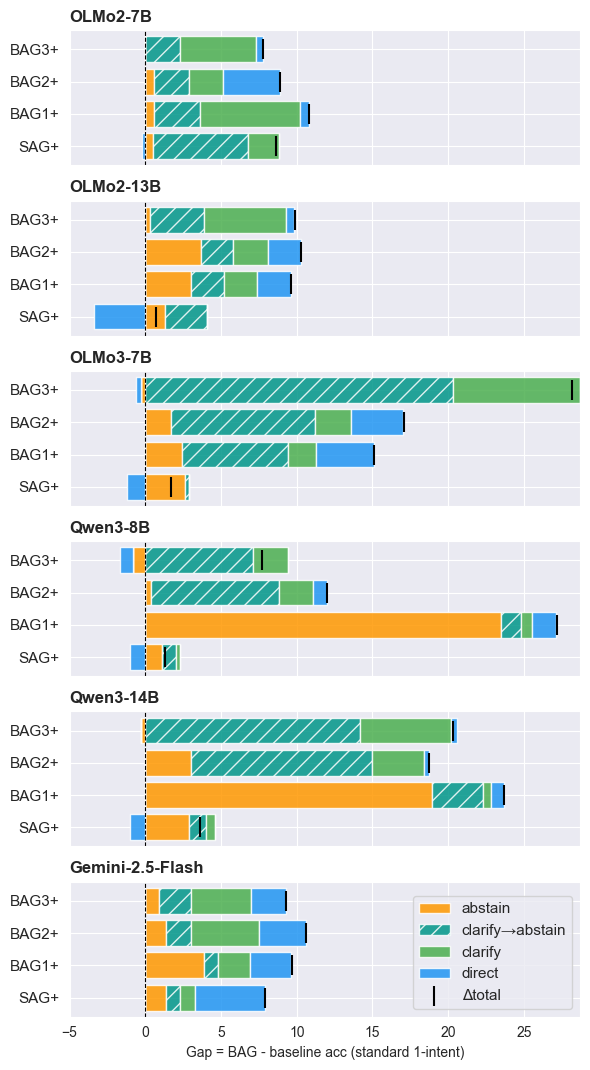}
    \caption{The contribution of each strategy to BAG+'s total accuracy increase (shown on the x-axis) for all models.}
    \label{fig:decomp_all_models}
\end{figure}

\begin{figure}
    \centering
    \includegraphics[width=0.99\linewidth]{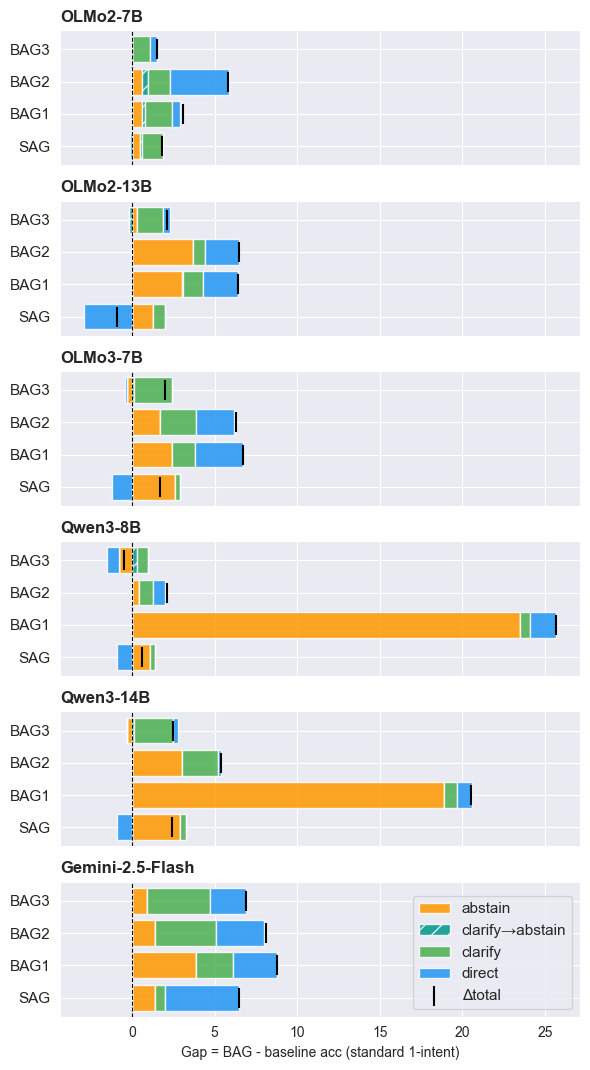}
    \caption{The contribution of each strategy to BAG's (so no second round of BAG for the final turn) total accuracy increase (shown on the x-axis) for all models.}
    \label{fig:decomp_all_nonplus}
\end{figure}

\input{figures/brevity_examples}

\input{figures/ambigqa_examples}

\section{Conversational Strategies without Augmentation}
\label{sec:analysis}

\begin{figure}
    \centering
    \includegraphics[width=1.0\linewidth]{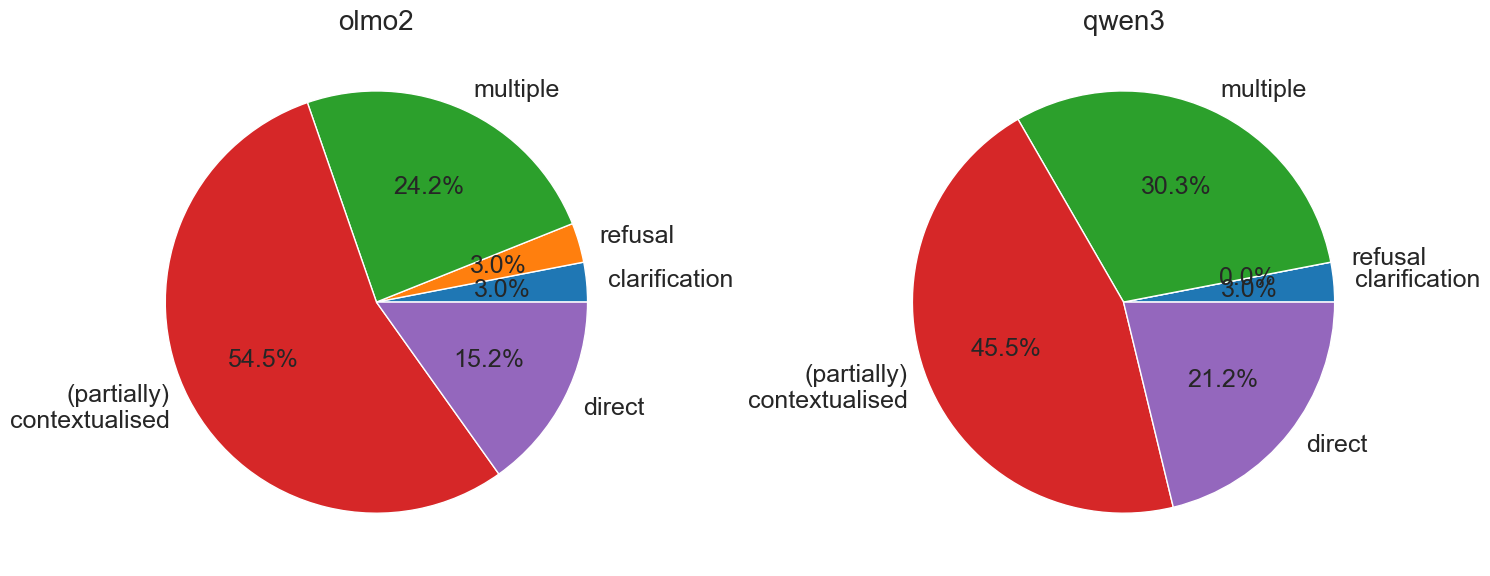}
    \caption{Strategy distribution for \textit{olmo2-7b-instruct} and \textit{qwen3-8b (non-thinking)} generations on 33 ambiguous AmbigQA questions with multiple intents. We manually annotated the generations with a single strategy each.}
    \label{fig:strategy_annotation}
\end{figure}

To get a sense of how standard models employ conversational actions, we select 33 random AmbigQA questions with multiple reference answers from the train set and inspect generations from the smallest \textsc{olmo2-7b-instruct} and \textsc{qwen3-8b} (non-thinking mode) models.

We annotate each generation with: \textbf{clarification question}, \textbf{refusal}, \textbf{multiple answers}, \textbf{(partially) contextualised answer}, or \textbf{direct answer} without any interpretation context. 

Figure \ref{fig:strategy_annotation} shows that models use a narrow range of conversational actions and rarely clarify or refuse. They mostly answer directly ($>94\%$), which we can further break down into (partially) contextualising their answer by assuming a specific interpretation of the question ($45-54\%$ \eg \textit{"In the US ..."}), giving multiple answers ($\sim24-30\%$), or answering directly without acknowledging ambiguity in any way ($15-20\%$).

\input{figures/brevity_results_table}


\begin{figure*}
    \centering
    \includegraphics[width=0.9\linewidth]{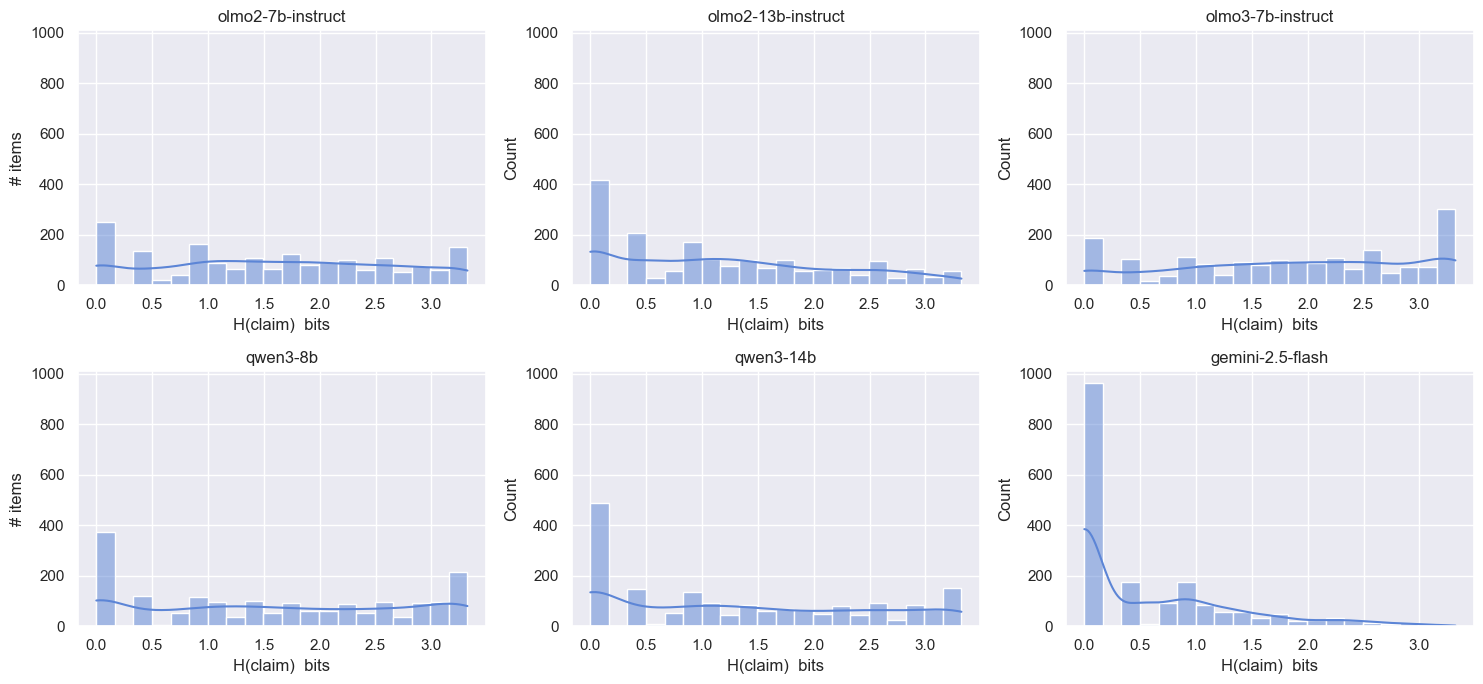}
    \caption{The distribution of belief state entropies across models. }
    \label{fig:entropy}
\end{figure*}

\begin{figure*}
    \centering
    \includegraphics[width=0.9\linewidth]{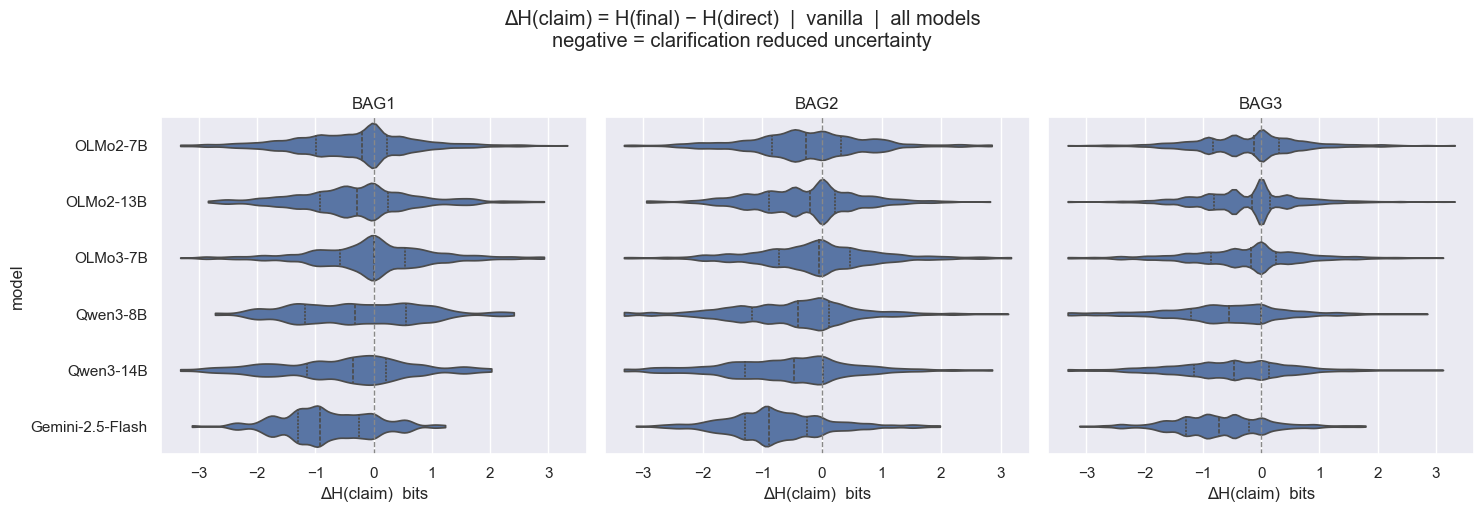}
    \caption{Uncertainty reduction after a clarification interaction for every model/prompt: largest for Gemini and smallest for Olmo3.}
    \label{fig:reduction}
\end{figure*}

%% file: figures/generation-settings.tex
\begin{figure*}[ht]

\begin{minipage}[t]{0.48\textwidth}
\noindent\small\textsc{(a) Direct generation standard baseline}
\begin{lstlisting}[style=promptstyle]
Who was the female singer on Gimme Shelter?
\end{lstlisting}
\end{minipage}
\hfill
\begin{minipage}[t]{0.48\textwidth}
\noindent\small\textsc{(b) Disambiguation upper bound}
\begin{lstlisting}[style=promptstyle]
Who was the female singer on the recorded version of Gimme Shelter?
\end{lstlisting}
\end{minipage}

\vspace{1em}

\begin{minipage}[t]{0.48\textwidth}
\noindent\small\textsc{(c) Strategy-Augmented Generation (SAG)}
\begin{lstlisting}[style=promptstyle]
# [1. Setting]
You are an AI assistant and your goal is to answer
a user's question directly, ask a clarification
question, or abstain.

# [3. Strategies \& Instructions]
- A direct answer is effective when the question is
  clear and you know the right answer.
- A clarification question is effective if it resolves
  ambiguity or underspecification in the user question,
  and if the user can answer it without needing to know
  the answer to their own question.
- Abstaining is effective if the question is clear but
  you don't know the right answer.

# [4. Context]
User question: Who was the female singer on Gimme Shelter?

# [5. Output format]
Format your response as:
STRATEGY: [DIRECT_ANSWER/CLARIFICATION_QUESTION/ABSTAIN]
REASONING: [...]
RESPONSE: [...]
\end{lstlisting}
\end{minipage}
\hfill
\begin{minipage}[t]{0.48\textwidth}
\noindent\small\textsc{(d) Belief-Augmented Generation (BAG)}
\begin{lstlisting}[style=promptstyle]
# [1. Setting] (same as SAG)
# [2. Belief state]
Below, you are given {K} candidate answers to the
user's question. Consider these your state of
knowledge and decide on your next action based on
the following criteria:

# [3. Strategy \& Instructions]
- A direct answer is effective when the question is
  clear and the candidate answers are consistent.
- A clarification question is effective if it resolves
  ambiguity or underspecification in the user question,
  and if the user can answer it without needing to know
  the answer to their own question.
- Abstaining is effective if the question is clear but
  the candidate answers are too contradictory to be
  reliable.

# [4. Context]
User question: Who was the female singer on Gimme Shelter?
Candidate answers: {A_1, A_2, ..., A_K}

# [5. Output format] (same as SAG)
\end{lstlisting}
\end{minipage}
\caption{The four generation settings. The direct generation baseline~(a) uses the original question. The disambiguation upper bound~(b) uses a disambiguation annotation. The SAG baseline~(c) prompts for conversational strategies without a belief state. \textsc{BAG}~(d) extends SAG by conditioning on $K$ generation samples. See Fig \ref{fig:prompt-belief6} and \ref{fig:prompt-belief7} for other two BAG variants.}
\label{fig:prompt-template}
\end{figure*}

%% file: figures/bag_prompts.tex


\begin{figure}[t]
\begin{lstlisting}[style=promptstyle]
You are a helpful AI assistant in conversation with a user that asked a question.

Below, you are given {len(belief_samples)} sampled generations from a language model given the user's question. Consider these your belief state - a representation of your uncertainty about the answer.

Your job is to analyze these generations and choose one of three strategies:

- DIRECT_ANSWER: The generations converge on the same answer - there is no meaningful uncertainty. Synthesize the most representative response from your belief state, at roughly the same length as the individual generations.
- CLARIFICATION_QUESTION: The generations reflect multiple semantically distinct but plausible answers, indicating the question is ambiguous or underspecified. Ask a clarification question that the user can answer by specifying their intent. Never ask the user to answer their own question.
- ABSTAIN: The generations are factually contradictory and inconsistent due to your lack of knowledge, AND no useful clarification question exists that would resolve the uncertainty. Politely decline to answer.

User question: {question}
Candidate answers:
{belief_state_text}

Format your output as:
STRATEGY: [DIRECT_ANSWER/CLARIFICATION_QUESTION/ABSTAIN]
REASONING: [A compact reasoning chain explaining your decision]
RESPONSE: [Your response to the user's question based on the (hidden) candidate answers, following your chosen strategy.]

Please provide your output.
\end{lstlisting}
\caption{Prompt variant \textsc{BAG2}.}
\label{fig:prompt-belief6}
\end{figure}

\begin{figure}[t]
\begin{lstlisting}[style=promptstyle]
You are a helpful AI assistant in conversation with a user.

Below are {K} candidate answers representing your belief state - your uncertainty about the answer to the user's question.

Analyze them in two steps, then choose a strategy and respond.

Step 1 - Cluster by meaning: Group the answers by what they assert. Ignore surface variation (wording, punctuation); group by the underlying fact or claim. Give each group a letter label and a representative answer.

Step 2 - Interpret each cluster: For each group, ask: "What interpretation of the user's question would naturally lead to this answer?" Write one interpretation per group. If you cannot identify any coherent interpretation for a group, mark it "uninterpretable."

Step 3 - Choose a strategy:
- DIRECT_ANSWER: there is only 1 meaningful cluster, or all clusters share a single interpretation. Give a complete, direct answer synthesized from the dominant cluster.
- CLARIFICATION_QUESTION: there are 2-3 clusters, each with a distinct coherent interpretation, and the user can specify their intent without already knowing the answer. Ask about the user's context or purpose - do NOT simply ask "do you mean X or Y?", instead ask about the situation or intent that would determine which interpretation applies.
- ABSTAIN: clusters are uninterpretable, too numerous to resolve, or no clarification question could realistically distinguish them. Politely decline.

User question: {question}
Candidate answers:
{belief_state_text}

Format your output as:
CLUSTERS: [one line per group, e.g. "A ({{k}}/{n}): <representative answer>"]
INTERPRETATIONS: [one line per group, e.g. "A -> <what interpretation of the question leads here>"]
STRATEGY: [DIRECT_ANSWER / CLARIFICATION_QUESTION / ABSTAIN]
REASONING: [one compact sentence explaining your routing decision]
RESPONSE: [your response to the user - they do not see the candidate answers, clusters, or reasoning]
\end{lstlisting}
\caption{Prompt variant \textsc{BAG3}.}
\label{fig:prompt-belief7}
\end{figure}

\begin{figure}[t]
\begin{lstlisting}[style=promptstyle]
Below are {n} candidate answers representing your belief state - independent attempts at answering the original question given our conversation so far.

Determine whether your belief state shows enough consensus to give a confident answer, or whether the candidates diverge too much (indicating you are guessing rather than knowing).

Step 1 - Assess consensus: Do the {n} candidate answers agree on the same core factual claim? Ignore minor surface differences in wording - focus on whether they assert the same fact. Count how many samples support the most common claim.

Step 2 - Route:
- DIRECT_ANSWER: The candidates largely agree (> 70% support the same claim). Synthesize a single confident answer based on the consensus.
- ABSTAIN: The candidates make multiple different factual claims with no clear majority. This means you are uncertain and should not guess. Politely decline.

Candidate answers:
{belief_state_text}

Format your output as:
CONSENSUS: [one line: "X/{n} candidates agree on: <the claim>" or "no consensus - candidates split across N different claims"]
STRATEGY: [DIRECT_ANSWER / ABSTAIN]
REASONING: [one compact sentence explaining your decision]
RESPONSE: [your answer to the user if DIRECT_ANSWER, or a polite decline if ABSTAIN]
\end{lstlisting}
\caption{Prompt for applying BAG a second time after the clarification interaction with the user, to determine whether to reply or abstain after all, because the dialogue history does not reduce uncertainty sufficiently. This prompt contains an example for how to use an explicit threshold in the prompt: if there are less than 7/10 consistent answers the instruction is to abstain.}
\label{fig:prompt-final_bag}
\end{figure}

%% file: figures/user_prompt.tex
  \begin{figure*}[t]
  \begin{minipage}[t]{0.48\linewidth}
  \begin{lstlisting}[style=promptstyle]
  Pretend you are roleplaying a user that
  asked a question to an AI assistant.

  The AI assistant asked a clarification
  question. Your task is to formulate the
  user's response to this clarification
  question given secret additional context.
  You will be given a disambiguated version
  of the user question to help resolve any
  ambiguity or underspecification.

  Rules:
  - Never reveal the final answer to the
    user question, only answer the AI's
    clarification question.
  - If the AI's clarification question is
    impossible to answer or not helpful,
    please respond "I don't know".

  User question (turn 1): {question}
  Secret disambiguated user question:
    {disambig_question}
  AI clarification question (turn 2):
    {clarification_question}
  User answer (turn 3): [your answer]

  Format your response as:
  REASONING: [A compact reasoning chain
  explaining your answer]
  USER ANSWER: [Your simulated user answer
  to the AI's clarification question]
  \end{lstlisting}
  \end{minipage}%
  \hfill%
  \begin{minipage}[t]{0.48\linewidth}
  \begin{lstlisting}[style=promptstyle]
  Pretend you are roleplaying a user that
  asked a question to an AI assistant.

  The AI assistant asked a clarification
  question. Your task is to formulate the
  user's response to this clarification
  question given secret additional context.
  You will be given the final reference
  answer to the user question to help
  resolve any ambiguity or underspecification.

  Rules:
  - Never reveal the final answer to the
    user question, only answer the AI's
    clarification question.
  - If the AI's clarification question is
    impossible to answer or not helpful,
    please respond "I don't know".

  User question (turn 1): {question}
  Secret final reference answer: {reference}
  AI clarification question (turn 2):
    {clarification_question}
  User answer (turn 3): [your answer]

  Format your response as:
  REASONING: [A compact reasoning chain
  explaining your answer]
  USER ANSWER: [Your simulated user answer
  to the AI's clarification question
  (DON'T MENTION {reference})]
  \end{lstlisting}
  \end{minipage}
  \caption{The two prompts for the user simulator LLM. The two prompts differ in the secret context provided: for ambiguous questions (left), we give a disambiguated version of the question; for non-ambiguous questions (right) we give the reference answer itself with an explicit instruction not to reveal it.}
  \label{fig:user-prompt}
  \end{figure*}

%% file: figures/judge_prompt.tex
  \begin{figure*}[t]
  \begin{minipage}[t]{0.48\linewidth}
  \begin{lstlisting}[style=promptstyle]
Your task is to evaluate whether a model answer correctly responds to a factual question.

Correct answer: {ref_text}

The correct answer above is ground truth - treat it as authoritative regardless of your own knowledge.

Guidelines:
- CORRECT ("yes"): the model directly asserts the correct answer as its main claim. Semantically equivalent forms are accepted (abbreviations, name variants, different date formats, singular/plural).
- INCORRECT ("no"): the model (a) states a different answer as its main claim, (b) mentions the correct answer only incidentally (e.g. as one item in a list, in a historical aside, or as a counterexample without endorsing it), or (c) is too vague or non-committal to constitute a clear answer.

Question: {question}
Model answer: {candidate}

Respond in this exact format:
REASONING: [one sentence explaining your decision]
VERDICT: yes / no
  \end{lstlisting}
  \end{minipage}%
  \hfill%
  \begin{minipage}[t]{0.48\linewidth}
  \begin{lstlisting}[style=promptstyle]
  Your task is to evaluate whether a model answer correctly responds to a factual question.

This question has multiple valid interpretations. The model answer is correct if it directly asserts ANY ONE of the following correct answers as its main claim:

{refs_block}

The correct answers above are ground truth - treat them as authoritative regardless of your own knowledge.

Guidelines:
- CORRECT ("yes"): the model directly asserts any one correct answer as its main claim. Semantically equivalent forms are accepted (abbreviations, name variants, different date formats, singular/plural).
- INCORRECT ("no"): the model (a) states a different answer as its main claim for none of the interpretations, (b) mentions a correct answer only incidentally (e.g. as one item in a list, in a historical aside, or as a counterexample without endorsing it), or (c) is too vague or non-committal to constitute a clear answer.

Question: {question}
Model answer: {candidate}

Respond in this exact format:
REASONING: [one sentence explaining your decision]
VERDICT: yes / no
  \end{lstlisting}
  \end{minipage}
  \caption{The two prompts for the LLM judge. On the left is the prompt to assess against a single user intent, the right assesses against any of the intents.}
  \label{fig:judge-prompt}
  \end{figure*}

%% file: figures/bag_results_nonplus.tex
\begin{table*}[ht]
\centering\small
\begin{tabular}{l|cc|cc|cccc}
\toprule
\multicolumn{1}{l|}{} & \multicolumn{4}{c|}{\textbf{Direct Generation}} & \multicolumn{4}{c}{\textbf{Augmented Generation}} \\
\cmidrule(lr){2-5} \cmidrule(lr){6-9}
 & \multicolumn{2}{c|}{1\,intent} & \multicolumn{2}{c|}{any\,intent} & \multicolumn{4}{c}{1\,intent} \\
\cmidrule(lr){2-3} \cmidrule(lr){4-5} \cmidrule(lr){6-9}
 & Standard & Disambig & Standard & Disambig & SAG & BAG1 & BAG2 & BAG3 \\
\midrule
OLMo2-7B & \cellcolor[rgb]{0.647,0.000,0.149}\textcolor{white}{33.9} & \cellcolor[rgb]{0.964,0.477,0.286}\textcolor{black}{36.3} & \cellcolor[rgb]{0.000,0.408,0.216}\textcolor{white}{44.8} & \cellcolor[rgb]{0.710,0.876,0.454}\textcolor{black}{41.2} & \cellcolor[rgb]{0.917,0.343,0.224}\textcolor{black}{33.5} & \cellcolor[rgb]{0.986,0.637,0.360}\textcolor{black}{37.0} & \cellcolor[rgb]{0.950,0.979,0.681}\textcolor{black}{39.7} & \cellcolor[rgb]{0.886,0.277,0.194}\textcolor{black}{35.4} \\
\arrayrulecolor{black!35}\specialrule{1pt}{2pt}{2pt}\arrayrulecolor{black}
OLMo2-13B & \cellcolor[rgb]{0.762,0.111,0.151}\textcolor{white}{45.3} & \cellcolor[rgb]{0.994,0.778,0.461}\textcolor{black}{49.6} & \cellcolor[rgb]{0.000,0.408,0.216}\textcolor{white}{59.2} & \cellcolor[rgb]{0.459,0.767,0.395}\textcolor{black}{55.9} & \cellcolor[rgb]{0.647,0.000,0.149}\textcolor{white}{44.6} & \cellcolor[rgb]{1.000,0.993,0.737}\textcolor{black}{51.7} & \cellcolor[rgb]{0.997,0.999,0.745}\textcolor{black}{51.8} & \cellcolor[rgb]{0.957,0.427,0.263}\textcolor{black}{47.4} \\
\arrayrulecolor{black!35}\specialrule{1pt}{2pt}{2pt}\arrayrulecolor{black}
OLMo3-7B & \cellcolor[rgb]{0.647,0.000,0.149}\textcolor{white}{24.7} & \cellcolor[rgb]{0.770,0.118,0.151}\textcolor{white}{25.2} & \cellcolor[rgb]{0.000,0.408,0.216}\textcolor{white}{32.4} & \cellcolor[rgb]{0.702,0.873,0.449}\textcolor{black}{29.9} & \cellcolor[rgb]{0.968,0.507,0.300}\textcolor{black}{29.6} & \cellcolor[rgb]{0.143,0.616,0.324}\textcolor{black}{31.5} & \cellcolor[rgb]{0.342,0.713,0.374}\textcolor{black}{31.0} & \cellcolor[rgb]{0.978,0.577,0.332}\textcolor{black}{26.7} \\
\arrayrulecolor{black!35}\specialrule{1pt}{2pt}{2pt}\arrayrulecolor{black}
Qwen3-8B & \cellcolor[rgb]{0.678,0.030,0.150}\textcolor{white}{33.1} & \cellcolor[rgb]{0.724,0.074,0.151}\textcolor{white}{33.7} & \cellcolor[rgb]{0.996,0.883,0.553}\textcolor{black}{43.2} & \cellcolor[rgb]{0.990,0.667,0.373}\textcolor{black}{40.3} & \cellcolor[rgb]{0.716,0.066,0.150}\textcolor{white}{33.4} & \cellcolor[rgb]{0.000,0.408,0.216}\textcolor{white}{58.8} & \cellcolor[rgb]{0.832,0.177,0.153}\textcolor{white}{35.1} & \cellcolor[rgb]{0.647,0.000,0.149}\textcolor{white}{32.6} \\
\arrayrulecolor{black!35}\specialrule{1pt}{2pt}{2pt}\arrayrulecolor{black}
Qwen3-14B & \cellcolor[rgb]{0.647,0.000,0.149}\textcolor{white}{40.1} & \cellcolor[rgb]{0.877,0.259,0.185}\textcolor{white}{42.8} & \cellcolor[rgb]{0.851,0.937,0.545}\textcolor{black}{52.4} & \cellcolor[rgb]{0.998,0.936,0.641}\textcolor{black}{49.3} & \cellcolor[rgb]{0.859,0.221,0.168}\textcolor{white}{41.8} & \cellcolor[rgb]{0.000,0.408,0.216}\textcolor{white}{60.6} & \cellcolor[rgb]{0.979,0.587,0.337}\textcolor{black}{45.5} & \cellcolor[rgb]{0.868,0.240,0.177}\textcolor{white}{42.6} \\
\arrayrulecolor{black!35}\specialrule{1pt}{2pt}{2pt}\arrayrulecolor{black}
Gemini-2.5-Flash & \cellcolor[rgb]{0.647,0.000,0.149}\textcolor{white}{59.1} & \cellcolor[rgb]{0.880,0.950,0.585}\textcolor{black}{68.7} & \cellcolor[rgb]{0.000,0.408,0.216}\textcolor{white}{75.6} & \cellcolor[rgb]{0.084,0.563,0.296}\textcolor{black}{74.2} & \cellcolor[rgb]{0.996,0.863,0.532}\textcolor{black}{66.0} & \cellcolor[rgb]{0.950,0.979,0.681}\textcolor{black}{67.9} & \cellcolor[rgb]{1.000,0.988,0.729}\textcolor{black}{67.2} & \cellcolor[rgb]{0.997,0.902,0.585}\textcolor{black}{66.0} \\
\bottomrule
\end{tabular}
\caption{Accuracy on AmbigQA for direct generation (left) and augmented generation applied only to the first assistant turn, i.e. a single round of BAG (right). BAG outperforms the standard baseline in 17/18 configurations, SAG in 15/18, and the disambiguation oracle in 11/18. Colours are row-normalised; see Table \ref{tab:bag_results} for BAG+.}
\label{tab:bagnonplus_results}
\end{table*}

%% file: figures/brevity_examples.tex
\begin{table*}[t]
\centering
\small
\begin{tabularx}{\textwidth}{p{3.2cm} X X}
\toprule
\textbf{Question} & \textbf{Verbose generation (excerpt)} & \textbf{Single-sentence generation} \\
\midrule

Which is the most populated country in Europe?
&
``\ldots Russia was Europe's most populous country, although it's important to note that \textbf{part of Russia lies within Asia}.\ldots''
&
``Russia is the most populous country in Europe.''
\\
\addlinespace

Who has the record for the most strikeouts?
&
``The \textbf{all-time} \textbf{Major League Baseball} record for the most strikeouts \textbf{by a pitcher} is held by Nolan Ryan with 5,714 strikeouts.\ldots''
&
``Nolan Ryan holds the MLB record for most strikeouts with 5,714.''
\\
\addlinespace

When do we celebrate Veterans Day this year?
&
``Veterans Day is observed on November 11 each year \textbf{in the United States}.\ldots''
&
``Veterans Day will be celebrated on November 11, 2023.''
\\
\addlinespace

When did Harry Potter and the Deathly Hallows Part 1 come out?
&
``\ldots was released on November 19, 2010, \textbf{in the United States}.\ldots''
&
``Harry Potter and the Deathly Hallows Part 1 was released on November 19, 2010.''
\\
\addlinespace

When did the first episode of Dragon Ball air?
&
``The first episode of `Dragon Ball' aired on February 26, 1986.\ldots''
&
``The first episode of `Dragon Ball' aired \textbf{in Japan} on February 26, 1986.''
\\

\bottomrule
\end{tabularx}
\caption{Effect of brevity-inducing prompts on ambiguity acknowledgement for \textit{olmo2-13b-instruct}. Bold text marks phrases that explicitly resolve an interpretation. When prompted for single-sentence answers, the model drops disambiguating context in four out of five cases, committing to an interpretation without signalling it.}
\label{table:examples_1sentence}
\end{table*}

%% file: figures/ambigqa_examples.tex
\begin{table*}[t]
\centering
\small
\setlength{\tabcolsep}{4pt}
\renewcommand{\arraystretch}{1.15}
\begin{tabular}{clp{5.0cm}p{5.8cm}}
\toprule
\textbf{\#} & \textbf{Type} & \textbf{Original question} & \textbf{Annotated disambiguations \& ref answers} \\
\midrule

1 & Genuine &
\textit{Who made the song ``These Boots Are Made for Walkin'\,'?} &
wrote $\to$ Lee Hazlewood \\
& & &
sang $\to$ Nancy Sinatra \\[3pt]

2 & Genuine &
\textit{Who the female singer on Gimme Shelter?} &
recorded version $\to$ Merry Clayton \\
& & &
on tour $\to$ Lisa Fischer \\[3pt]

\midrule

3 & Time-sensitive &
\textit{Who is the \textbf{current} chairman of the African Union Commission?} &
became chairman in 2017 $\to$ Moussa Faki \\
& & &
became chairman in 2012 $\to$ Nkosazana Dlamini-Zuma \\
& & &
became chairman in 2008 $\to$ Jean Ping \\[3pt]

4 & Time-sensitive &
\textit{When did the \textbf{current} queen of England take the throne?} &
[single answer] $\to$ 6 February 1952 \textit{(outdated since 2022)} \\[3pt]

\midrule

5 & Missing ambiguity &
\textit{When was the movie Rudolph the red-nosed reindier made} &
original short $\to$ 1948 \\
& & &
remake $\to$ 1998 \textit{(misses iconic 1964 Rankin/Bass TV special)}  \\[3pt]

6 & Missing ambiguity &
\textit{What channel is the new show FBI on?} &
[single answer] $\to$ CBS \textit{(US-centric; varies by country)} \\[3pt]

\midrule

7 & Granularity &
\textit{`In which episode of Smallville does Jonathan die?} &
title $\to$ reckoning \\
& & &
number $\to$ season 5 episode 12 \\[3pt]

8 & Granularity &
\textit{What do the symbols on the Indian flag represent?} &
the wheel $\to$ Dharma Chakra \\
& & &
the saffron stripe $\to$ courage and sacrifice \\
& & &
the white stripe $\to$ peace and truth \\
& & &
the green stripe $\to$ faith and chivalry \\
& & &
the Calcutta flag lotuses $\to$ the eight provinces \\[3pt]

\midrule

9 & Marginal interp. &
\textit{How long do contestants get to answer on Jeopardy?} &
regular question $\to$ 5 seconds \\
& & &
Final Jeopardy $\to$ 30 seconds \\
& & &
\textit{online contestant test} $\to$ 15 seconds \\

10 & Marginal interp. &
\textit{What is the population of Venice Italy 2018?} &
total population $\to$ 260.897 \\
& & &
population density $\to$ 1,600/sq mi \\

\bottomrule
\end{tabular}
\caption{
  AmbigQA examples.
  Examples~1--2 show genuine ambiguity well suited to clarification.
  Examples~3--4 show time sensitivity questions that are annotated as ambiguity, and can also become outdated.
  Example~5--6 shows ambiguity the dataset \emph{misses}, \eg due to US-centric annotations.
  Examples~7--8 show overly granular disambiguations.
  Example~9--10 includes a marginal interpretation alongside a very salient one.
}
\label{tab:ambigqa-examples}
\end{table*}

%% file: figures/brevity_results_table.tex
\begin{table*}[ht]
\centering\scriptsize
\begin{tabular}{ll|ccc|cc|cccc}
\toprule
\multicolumn{2}{l|}{} & \multicolumn{5}{c|}{\textbf{Direct Generation}} & \multicolumn{4}{c}{\textbf{Augmented Generation}} \\
\cmidrule(lr){3-7} \cmidrule(lr){8-11}
 &  & \multicolumn{3}{c|}{1\,intent} & \multicolumn{2}{c|}{any\,intent} & \multicolumn{4}{c}{1\,intent} \\
\cmidrule(lr){3-5} \cmidrule(lr){6-7} \cmidrule(lr){8-11}
 &  & Standard & Disambig & Standard & Disambig & words & SAG+ & BAG1+ & BAG2+ & BAG3+ \\
\midrule
\multirow{3}{*}{OLMo2-7B} & free & \cellcolor[rgb]{0.647,0.000,0.149}\textcolor{white}{33.9} & \cellcolor[rgb]{0.964,0.477,0.286}\textcolor{black}{36.3} & \cellcolor[rgb]{0.000,0.408,0.216}\textcolor{white}{44.8} & \cellcolor[rgb]{0.710,0.876,0.454}\textcolor{black}{41.2} & 149.9 & \cellcolor[rgb]{0.655,0.853,0.418}\textcolor{black}{41.5} & \cellcolor[rgb]{0.008,0.423,0.223}\textcolor{white}{44.7} & \cellcolor[rgb]{0.342,0.713,0.374}\textcolor{black}{42.8} & \cellcolor[rgb]{0.607,0.832,0.411}\textcolor{black}{41.7} \\
\cline{2-11}
 & concise & \cellcolor[rgb]{0.647,0.000,0.149}\textcolor{white}{31.9} & \cellcolor[rgb]{0.961,0.457,0.277}\textcolor{black}{34.7} & \cellcolor[rgb]{0.388,0.735,0.385}\textcolor{black}{42.5} & \cellcolor[rgb]{0.765,0.900,0.489}\textcolor{black}{40.4} & 42.9 & \cellcolor[rgb]{0.577,0.819,0.408}\textcolor{black}{41.5} & \cellcolor[rgb]{0.000,0.408,0.216}\textcolor{white}{45.1} & \cellcolor[rgb]{0.044,0.489,0.258}\textcolor{white}{44.5} & \cellcolor[rgb]{0.119,0.605,0.318}\textcolor{black}{43.7} \\
\cline{2-11}
 & sentence & \cellcolor[rgb]{0.647,0.000,0.149}\textcolor{white}{31.8} & \cellcolor[rgb]{0.894,0.296,0.202}\textcolor{black}{34.1} & \cellcolor[rgb]{0.796,0.914,0.510}\textcolor{black}{41.6} & \cellcolor[rgb]{0.999,0.969,0.697}\textcolor{black}{39.2} & 13.5 & \cellcolor[rgb]{0.804,0.917,0.515}\textcolor{black}{41.5} & \cellcolor[rgb]{0.655,0.853,0.418}\textcolor{black}{42.7} & \cellcolor[rgb]{0.000,0.408,0.216}\textcolor{white}{47.4} & \cellcolor[rgb]{0.636,0.845,0.414}\textcolor{black}{42.8} \\
\arrayrulecolor{black!35}\specialrule{1pt}{2pt}{2pt}\arrayrulecolor{black}
\multirow{3}{*}{OLMo2-13B} & free & \cellcolor[rgb]{0.647,0.000,0.149}\textcolor{white}{45.3} & \cellcolor[rgb]{0.993,0.702,0.397}\textcolor{black}{49.6} & \cellcolor[rgb]{0.000,0.408,0.216}\textcolor{white}{59.2} & \cellcolor[rgb]{0.489,0.780,0.398}\textcolor{black}{55.9} & 146.3 & \cellcolor[rgb]{0.850,0.202,0.159}\textcolor{white}{46.8} & \cellcolor[rgb]{0.671,0.859,0.428}\textcolor{black}{54.9} & \cellcolor[rgb]{0.548,0.806,0.404}\textcolor{black}{55.6} & \cellcolor[rgb]{0.636,0.845,0.414}\textcolor{black}{55.1} \\
\cline{2-11}
 & concise & \cellcolor[rgb]{0.647,0.000,0.149}\textcolor{white}{43.4} & \cellcolor[rgb]{0.943,0.399,0.250}\textcolor{black}{45.9} & \cellcolor[rgb]{0.000,0.408,0.216}\textcolor{white}{56.7} & \cellcolor[rgb]{0.498,0.784,0.399}\textcolor{black}{53.5} & 53.0 & \cellcolor[rgb]{0.976,0.567,0.327}\textcolor{black}{46.8} & \cellcolor[rgb]{0.459,0.767,0.395}\textcolor{black}{53.7} & \cellcolor[rgb]{0.068,0.533,0.281}\textcolor{black}{55.8} & \cellcolor[rgb]{0.178,0.633,0.333}\textcolor{black}{55.0} \\
\cline{2-11}
 & sentence & \cellcolor[rgb]{0.647,0.000,0.149}\textcolor{white}{42.3} & \cellcolor[rgb]{0.990,0.667,0.373}\textcolor{black}{46.3} & \cellcolor[rgb]{0.028,0.460,0.243}\textcolor{white}{55.4} & \cellcolor[rgb]{0.353,0.718,0.377}\textcolor{black}{53.3} & 14.6 & \cellcolor[rgb]{0.993,0.748,0.435}\textcolor{black}{46.8} & \cellcolor[rgb]{0.201,0.644,0.339}\textcolor{black}{54.0} & \cellcolor[rgb]{0.330,0.707,0.371}\textcolor{black}{53.4} & \cellcolor[rgb]{0.000,0.408,0.216}\textcolor{white}{55.8} \\
\arrayrulecolor{black!35}\specialrule{1pt}{2pt}{2pt}\arrayrulecolor{black}
\multirow{3}{*}{OLMo3-7B} & free & \cellcolor[rgb]{0.647,0.000,0.149}\textcolor{white}{24.7} & \cellcolor[rgb]{0.678,0.030,0.150}\textcolor{white}{25.2} & \cellcolor[rgb]{0.982,0.607,0.346}\textcolor{black}{32.4} & \cellcolor[rgb]{0.939,0.390,0.246}\textcolor{black}{29.9} & 125.5 & \cellcolor[rgb]{0.921,0.352,0.228}\textcolor{black}{29.5} & \cellcolor[rgb]{0.944,0.977,0.673}\textcolor{black}{39.8} & \cellcolor[rgb]{0.835,0.930,0.535}\textcolor{black}{41.8} & \cellcolor[rgb]{0.000,0.408,0.216}\textcolor{white}{52.9} \\
\cline{2-11}
 & concise & \cellcolor[rgb]{0.647,0.000,0.149}\textcolor{white}{26.5} & \cellcolor[rgb]{0.678,0.030,0.150}\textcolor{white}{27.0} & \cellcolor[rgb]{0.993,0.717,0.409}\textcolor{black}{34.9} & \cellcolor[rgb]{0.952,0.418,0.258}\textcolor{black}{31.7} & 35.3 & \cellcolor[rgb]{0.859,0.221,0.168}\textcolor{white}{29.5} & \cellcolor[rgb]{0.479,0.776,0.397}\textcolor{black}{46.8} & \cellcolor[rgb]{0.048,0.496,0.262}\textcolor{white}{51.6} & \cellcolor[rgb]{0.000,0.408,0.216}\textcolor{white}{52.9} \\
\cline{2-11}
 & sentence & \cellcolor[rgb]{0.647,0.000,0.149}\textcolor{white}{24.5} & \cellcolor[rgb]{0.755,0.103,0.151}\textcolor{white}{26.0} & \cellcolor[rgb]{0.993,0.717,0.409}\textcolor{black}{33.0} & \cellcolor[rgb]{0.976,0.567,0.327}\textcolor{black}{31.3} & 14.0 & \cellcolor[rgb]{0.943,0.399,0.250}\textcolor{black}{29.5} & \cellcolor[rgb]{0.587,0.823,0.409}\textcolor{black}{43.8} & \cellcolor[rgb]{0.000,0.408,0.216}\textcolor{white}{51.1} & \cellcolor[rgb]{0.044,0.489,0.258}\textcolor{white}{49.9} \\
\arrayrulecolor{black!35}\specialrule{1pt}{2pt}{2pt}\arrayrulecolor{black}
\multirow{3}{*}{Qwen3-8B} & free & \cellcolor[rgb]{0.647,0.000,0.149}\textcolor{white}{33.1} & \cellcolor[rgb]{0.686,0.037,0.150}\textcolor{white}{33.7} & \cellcolor[rgb]{0.995,0.825,0.500}\textcolor{black}{43.2} & \cellcolor[rgb]{0.980,0.597,0.341}\textcolor{black}{40.3} & 111.0 & \cellcolor[rgb]{0.755,0.103,0.151}\textcolor{white}{34.6} & \cellcolor[rgb]{0.000,0.408,0.216}\textcolor{white}{60.2} & \cellcolor[rgb]{0.998,0.926,0.625}\textcolor{black}{45.0} & \cellcolor[rgb]{0.986,0.637,0.360}\textcolor{black}{40.8} \\
\cline{2-11}
 & concise & \cellcolor[rgb]{0.647,0.000,0.149}\textcolor{white}{32.0} & \cellcolor[rgb]{0.877,0.259,0.185}\textcolor{white}{34.3} & \cellcolor[rgb]{0.939,0.974,0.665}\textcolor{black}{41.6} & \cellcolor[rgb]{0.999,0.974,0.705}\textcolor{black}{40.5} & 18.3 & \cellcolor[rgb]{0.894,0.296,0.202}\textcolor{black}{34.6} & \cellcolor[rgb]{0.000,0.408,0.216}\textcolor{white}{49.7} & \cellcolor[rgb]{0.788,0.910,0.504}\textcolor{black}{43.2} & \cellcolor[rgb]{0.996,0.888,0.561}\textcolor{black}{39.2} \\
\cline{2-11}
 & sentence & \cellcolor[rgb]{0.647,0.000,0.149}\textcolor{white}{30.6} & \cellcolor[rgb]{0.948,0.409,0.254}\textcolor{black}{33.6} & \cellcolor[rgb]{0.773,0.903,0.494}\textcolor{black}{40.5} & \cellcolor[rgb]{0.985,0.994,0.729}\textcolor{black}{38.5} & 13.3 & \cellcolor[rgb]{0.978,0.577,0.332}\textcolor{black}{34.6} & \cellcolor[rgb]{0.000,0.408,0.216}\textcolor{white}{46.1} & \cellcolor[rgb]{0.788,0.910,0.504}\textcolor{black}{40.4} & \cellcolor[rgb]{0.956,0.982,0.689}\textcolor{black}{38.8} \\
\arrayrulecolor{black!35}\specialrule{1pt}{2pt}{2pt}\arrayrulecolor{black}
\multirow{3}{*}{Qwen3-14B} & free & \cellcolor[rgb]{0.647,0.000,0.149}\textcolor{white}{40.1} & \cellcolor[rgb]{0.859,0.221,0.168}\textcolor{white}{42.8} & \cellcolor[rgb]{0.974,0.989,0.713}\textcolor{black}{52.4} & \cellcolor[rgb]{0.996,0.855,0.526}\textcolor{black}{49.3} & 99.3 & \cellcolor[rgb]{0.859,0.221,0.168}\textcolor{white}{42.8} & \cellcolor[rgb]{0.000,0.408,0.216}\textcolor{white}{63.8} & \cellcolor[rgb]{0.430,0.754,0.391}\textcolor{black}{58.8} & \cellcolor[rgb]{0.225,0.656,0.344}\textcolor{black}{60.4} \\
\cline{2-11}
 & concise & \cellcolor[rgb]{0.647,0.000,0.149}\textcolor{white}{39.6} & \cellcolor[rgb]{0.881,0.268,0.190}\textcolor{white}{42.4} & \cellcolor[rgb]{0.857,0.940,0.553}\textcolor{black}{52.0} & \cellcolor[rgb]{0.999,0.979,0.713}\textcolor{black}{49.6} & 23.6 & \cellcolor[rgb]{0.903,0.315,0.211}\textcolor{black}{42.8} & \cellcolor[rgb]{0.000,0.408,0.216}\textcolor{white}{60.4} & \cellcolor[rgb]{0.056,0.511,0.270}\textcolor{white}{59.2} & \cellcolor[rgb]{0.260,0.673,0.353}\textcolor{black}{57.2} \\
\cline{2-11}
 & sentence & \cellcolor[rgb]{0.647,0.000,0.149}\textcolor{white}{37.5} & \cellcolor[rgb]{0.969,0.517,0.304}\textcolor{black}{41.6} & \cellcolor[rgb]{0.636,0.845,0.414}\textcolor{black}{49.8} & \cellcolor[rgb]{0.796,0.914,0.510}\textcolor{black}{48.4} & 14.7 & \cellcolor[rgb]{0.992,0.686,0.384}\textcolor{black}{42.8} & \cellcolor[rgb]{0.190,0.639,0.336}\textcolor{black}{52.6} & \cellcolor[rgb]{0.000,0.408,0.216}\textcolor{white}{54.9} & \cellcolor[rgb]{0.004,0.415,0.220}\textcolor{white}{54.8} \\
\arrayrulecolor{black!35}\specialrule{1pt}{2pt}{2pt}\arrayrulecolor{black}
\multirow{3}{*}{Gemini-2.5-Flash} & free & \cellcolor[rgb]{0.647,0.000,0.149}\textcolor{white}{59.1} & \cellcolor[rgb]{0.880,0.950,0.585}\textcolor{black}{68.7} & \cellcolor[rgb]{0.000,0.408,0.216}\textcolor{white}{75.6} & \cellcolor[rgb]{0.084,0.563,0.296}\textcolor{black}{74.2} & 46.5 & \cellcolor[rgb]{0.997,0.999,0.745}\textcolor{black}{67.4} & \cellcolor[rgb]{0.869,0.945,0.569}\textcolor{black}{68.8} & \cellcolor[rgb]{0.749,0.893,0.479}\textcolor{black}{69.8} & \cellcolor[rgb]{0.904,0.959,0.617}\textcolor{black}{68.4} \\
\cline{2-11}
 & concise & \cellcolor[rgb]{0.647,0.000,0.149}\textcolor{white}{52.4} & \cellcolor[rgb]{0.863,0.942,0.561}\textcolor{black}{61.6} & \cellcolor[rgb]{0.000,0.408,0.216}\textcolor{white}{67.9} & \cellcolor[rgb]{0.072,0.541,0.285}\textcolor{black}{66.8} & 5.9 & \cellcolor[rgb]{0.032,0.467,0.246}\textcolor{white}{67.4} & \cellcolor[rgb]{0.655,0.853,0.418}\textcolor{black}{63.2} & \cellcolor[rgb]{0.548,0.806,0.404}\textcolor{black}{63.9} & \cellcolor[rgb]{0.678,0.863,0.433}\textcolor{black}{63.0} \\
\cline{2-11}
 & sentence & \cellcolor[rgb]{0.647,0.000,0.149}\textcolor{white}{55.6} & \cellcolor[rgb]{0.710,0.876,0.454}\textcolor{black}{66.2} & \cellcolor[rgb]{0.000,0.408,0.216}\textcolor{white}{71.4} & \cellcolor[rgb]{0.064,0.526,0.277}\textcolor{black}{70.4} & 12.7 & \cellcolor[rgb]{0.528,0.797,0.402}\textcolor{black}{67.4} & \cellcolor[rgb]{0.597,0.827,0.410}\textcolor{black}{67.0} & \cellcolor[rgb]{0.796,0.914,0.510}\textcolor{black}{65.5} & \cellcolor[rgb]{0.773,0.903,0.494}\textcolor{black}{65.7} \\
\bottomrule
\end{tabular}
\caption{QA performance on 1832 AmbigQA-dev questions for all brevity-inducing settings. Colours per-row normalized. We can see that direct generation accuracy is not greatly impacted, but that the average direct generation length is, and therefore the belief state, with about a factor 10 for the open models from no (free) to sentence brevity. However, brevity-inducing prompts do affect BAG+ accuracy. There may be a trade-off between the size and complexity of belief states, and corresponding difficulties of analysing such large context windows, especially for smaller older models like Olmo2-7B, whereas, bigger models may more easily leverage the extra context.}
\label{tab:bag_brevity_results}
\end{table*}